\documentclass[10pt,twocolumn,letterpaper]{article}

\usepackage{cvpr}              

\usepackage{graphicx}
\usepackage{amsmath}
\usepackage{amssymb}
\usepackage{booktabs}

\usepackage{enumitem}
\usepackage[pagebackref,breaklinks,colorlinks]{hyperref}
\usepackage{bbding}

\usepackage[capitalize]{cleveref}
\crefname{section}{Sec.}{Secs.}
\Crefname{section}{Section}{Sections}
\Crefname{table}{Table}{Tables}
\crefname{table}{Tab.}{Tabs.}

\def\cvprPaperID{0000} 
\def\confName{CVPR}
\def\confYear{2022}

\begin{document}

\title{High-Fidelity GAN Inversion for Image Attribute Editing}

\author{Tengfei Wang$^1$ \quad
Yong Zhang$^2$\   \quad
Yanbo Fan$^2$  \quad
Jue Wang$^2$  \quad
Qifeng Chen$^1$ \Envelope
\\
\vspace{2mm}
{$^1$The Hong Kong University of Science and Technology \quad $^2$Tencent AI Lab} 
}

\maketitle

\begin{abstract}
We present a novel high-fidelity generative adversarial network (GAN) inversion framework that enables attribute editing with image-specific details well-preserved (e.g., background, appearance, and illumination). We first analyze the challenges of high-fidelity GAN inversion from the perspective of lossy data compression.  With a low bit-rate latent code, previous works have difficulties in preserving high-fidelity details in reconstructed and edited images. Increasing the size of a latent code can improve the accuracy of GAN inversion but at the cost of inferior editability. To improve image fidelity without compromising editability, we propose a distortion consultation approach that employs a distortion map as a reference for  high-fidelity reconstruction. In the distortion consultation inversion (DCI), the distortion map is first projected to a high-rate latent map, which then complements the basic low-rate latent code with more details via consultation fusion. To achieve high-fidelity editing, we propose an adaptive distortion alignment (ADA) module with a self-supervised training scheme, which bridges the gap between the edited and inversion images. Extensive experiments in the face and car domains show a clear improvement in both inversion and editing quality. The project page is \url{https://tengfei-wang.github.io/HFGI/}.
\end{abstract}

\vspace{-2mm}
\section{Introduction} 
Image attribute editing is the task of modifying desired attributes of a given image while preserving other details. With the rapid advancement of generative adversarial networks (GANs)~\cite{goodfellow2014generative},  a promising direction is to manipulate images with the strong control capacity of StyleGAN~\cite{karras2019style,karras2020analyzing}. To enable real-world image editing, GAN inversion techniques~\cite{xia2021gan} have been recently explored, which aim at projecting images to the latent space of a pre-trained GAN generator.

\begin{figure}[t]
    \centering 
    \footnotesize
    \begin{tabular}{@{}c@{\hspace{0.1mm}}c@{\hspace{0.3mm}}c@{\hspace{0.2mm}}c@{\hspace{0.2mm}}c@{\hspace{0.2mm}}c@{}}
    
      \rotatebox{90}{\hspace{4mm} Input}&
      &
     \includegraphics[width=0.198\linewidth]{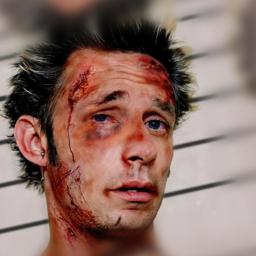}&   
     \includegraphics[width=0.198\linewidth]{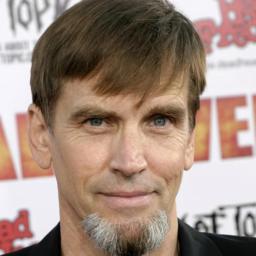}&  
    \includegraphics[width=0.265\linewidth]{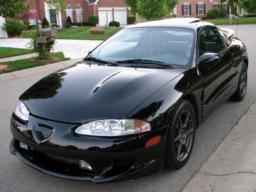}& 
     \includegraphics[width=0.265\linewidth]{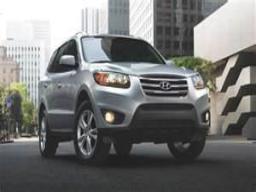}\\      
     
      \rotatebox{90}{\hspace{2mm} Inversion} &
      \rotatebox{90}{ \hspace{4mm} (e4e)} & 
     \includegraphics[width=0.199\linewidth]{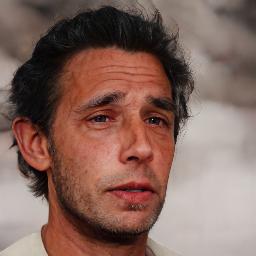}&   
     \includegraphics[width=0.199\linewidth]{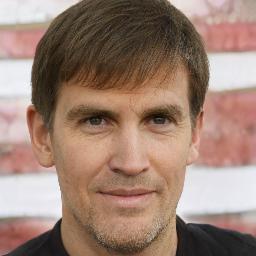}&  
     \includegraphics[width=0.265\linewidth]{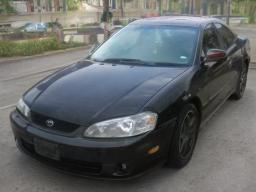}&   
    \includegraphics[width=0.265\linewidth]{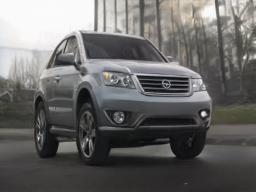}\\  
    
       \rotatebox{90}{\hspace{2mm} Inversion} &
       \rotatebox{90}{\hspace{2mm} (Restyle)}& 
     \includegraphics[width=0.199\linewidth]{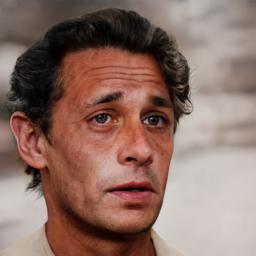}&  \includegraphics[width=0.199\linewidth]{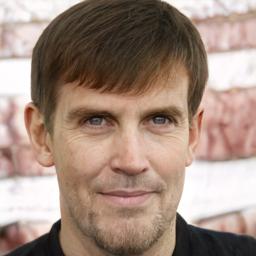}&
      \includegraphics[width=0.265\linewidth]{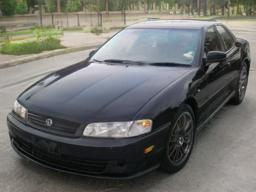}&   
      \includegraphics[width=0.265\linewidth]{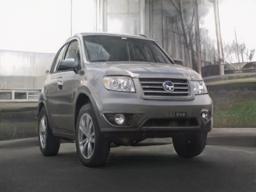}\\  
      
       \rotatebox{90}{\hspace{2mm} Inversion} &
       \rotatebox{90}{\hspace{4mm} (Ours)}& 
     \includegraphics[width=0.199\linewidth]{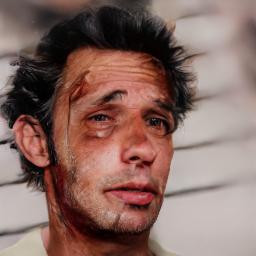}&  
\includegraphics[width=0.199\linewidth]{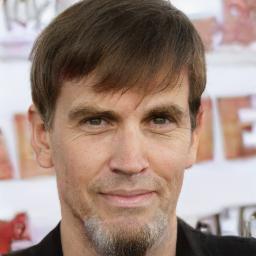}& 
      \includegraphics[width=0.265\linewidth]{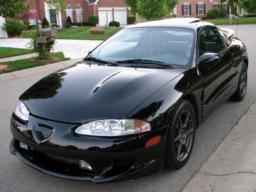}&     
       \includegraphics[width=0.265\linewidth]{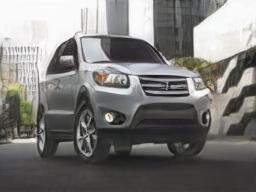}\\   
       
      \rotatebox{90}{\hspace{4mm} Edit 1} & 
      \rotatebox{90}{\hspace{4mm} (Ours)} &
     \includegraphics[width=0.199\linewidth]{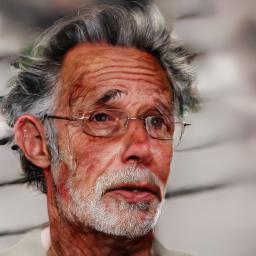}&
  \includegraphics[width=0.199\linewidth]{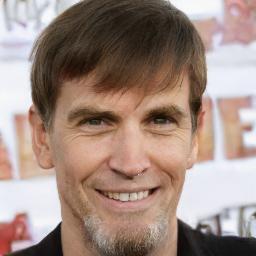}&    
     \includegraphics[width=0.265\linewidth]{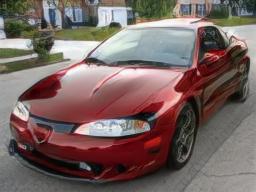}&  
      \includegraphics[width=0.265\linewidth]{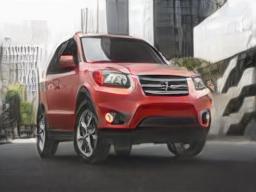}\\ 
      
     \rotatebox{90}{\hspace{4mm} Edit 2} &  
     \rotatebox{90}{\hspace{4mm} (Ours)}&
     \includegraphics[width=0.199\linewidth]{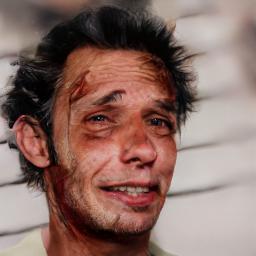} &
          \includegraphics[width=0.199\linewidth]{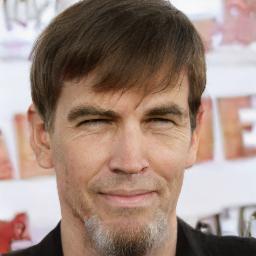} &
     \includegraphics[width=0.265\linewidth]{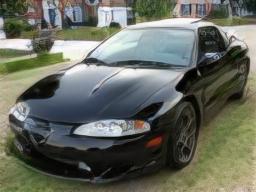} &
     \includegraphics[width=0.265\linewidth]{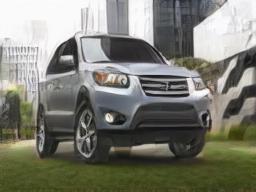} \\     

    \end{tabular}
    \vspace{-3mm}
   \caption{High-fidelity image inversion and editing (age, smile, eyes, color, grass). Our method  performs well on details preservation in both inverted and edited results such as background, makeup, beard/hair style, reflection and shadow.}
    \label{fig:teaser}
    \vspace{-4mm}
\end{figure}

Existing GAN inversion approaches either perform  per-image optimization~\cite{zhu2016generative,abdal2019image2stylegan,kang2021gan} or learn a data-driven encoder~\cite{richardson2020encoding,tov2021designing}. Optimization approaches achieve higher reconstruction accuracy by over-fitting on a single image, but the latent code may get out of GAN manifold, leading to inferior editing
quality. In contrast, encoder-based GAN inversion methods are faster  and show better editing performance due to knowledge learned from numerous training images. Nevertheless, their reconstruction results are usually inaccurate and of low fidelity: these methods can reconstruct a coarse layout (low-frequency patterns), but the image-specific details (high-frequency patterns) are often ignored. For example, the reconstructed face images typically possess averaged patterns that agree with the majority of training images (e.g., normal pose/expression, occlusion/shadow-free), and the details that present minority patterns (e.g., background, illumination, accessory) in training data are subject to distortion. It is highly desirable to preserve these image-specific details in reconstruction and editing  with high fidelity.

Though some  works tried to improve the reconstruction accuracy of encoder-based methods, their editing performance usually decreases~\cite{tov2021designing}.  To  analyze the limitation of existing approaches, we consider the GAN inversion problem as a lossy data compression system with a frozen decoder. According to Rate-Distortion theory~\cite{shannon1959coding}, reversing a real-world image to a low-dimensional latent code would inevitably lead to information loss. As conjectured by information bottleneck theory~\cite{tishby2015deep}, the lost information is primarily image-specific details as the deep compression model tends to retain common information of a domain. Based on these analyses and experimental observations, we present the  Rate-Distortion-Edit trade-off  for GAN inversion, which further inspires our framework.

According to this trade-off, the low-rate latent codes are insufficient for high-fidelity GAN inversion. However, it is non-trivial to improve the reconstruction accuracy by directly increasing the rate. A higher-rate latent codes can easily achieve a low distortion  by overfiting on the reconstruction process, but  would suffer a dramatic editing performance drop. To achieve both accuracy and editability (high-fidelity editing), we propose a novel framework that equips low-rate encoder  models with  \textbf{distortion consultation}. The consultation branch serves as a `cheat sheet'  for generation that only conveys the ignored image-specific information. Specifically, we leverage the distortion map between source and low-fidelity reconstructed image as a reference and project it to  higher-rate latent maps. Compared with high-rate latent codes  inferred from a full image, the distortion map only conveys image-specific details and can thus alleviate the aforementioned overfitting issue. The high-rate latent map and low-rate latent code are further embedded and fused in the generator via \textbf{consultation fusion}. Our scheme shows a clear improvement in reconstruction quality, and no test-time optimization is involved.

For attribute editing, following previous works, we perform vector arithmetic~\cite{radford2015unsupervised} on the low-rate latent code, while the consultation  is desired to bring back lost details. While the distortion consultation substantially contributes to the inversion quality, it cannot directly apply the  distortion map observed on the inversion image for editing due to the misalignment between inverted and edited images. To this end, we additionally design an \textbf{adaptive distortion alignment (ADA)} network to adjust the distortion map with the edited images. To disentangle the alignment from the consultation encoder and stabilize the training, we impose intermediate supervision on ADA by proposing   an alignment regularization with a self-supervised training scheme. 

Extensive experiments show that our method significantly outperforms current  approaches in terms of details preservation in both reconstructed and edited results. On account of the high-fidelity inversion capacity, our approach is robust to viewpoint and illumination fluctuation and can thus perform temporally consistent editing on  videos. Our primary contributions can be summarized as follows.

\begin{itemize}[noitemsep,topsep=0pt]
  \item We propose a distortion consultation inversion scheme that combines both high reconstruction quality and compelling editability with consultation  fusion.  
  
  \item For high-fidelity  editing, we propose the  adaptive distortion alignment module with a self-supervised learning scheme. By alignment, the distortion information can be propagated well to the edited images.
  
  \item Our method outperforms state-of-the-art approaches qualitatively and quantitatively on diverse image domains and videos. The framework is simple, fast and can be easily applied to GAN models. 
\end{itemize}

\begin{figure*}[t]
    \centering 
    \footnotesize
    \includegraphics[width=0.95 \linewidth]{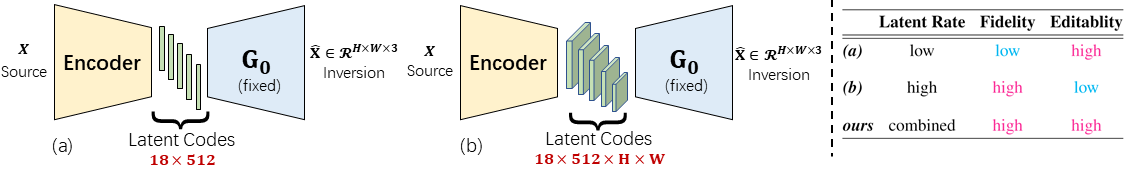}\\
    
    \begin{tabular}{@{}c@{\hspace{0.2mm}}c@{\hspace{0.2mm}}c@{\hspace{0.2mm}}c@{\hspace{0.2mm}}c@{\hspace{0.2mm}}c@{\hspace{0.2mm}}c@{\hspace{0.2mm}}c@{}}
    \rotatebox{90}{\hspace{5mm }+ Beard} &
     \includegraphics[width=0.132\linewidth]{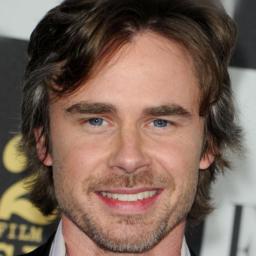}  & 
     \includegraphics[width=0.132\linewidth]{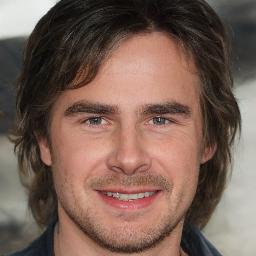}&  
     \includegraphics[width=0.132\linewidth]{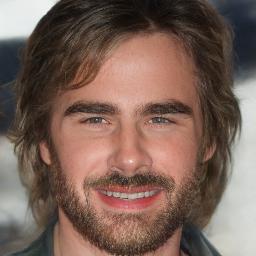} &  
     \includegraphics[width=0.132\linewidth]{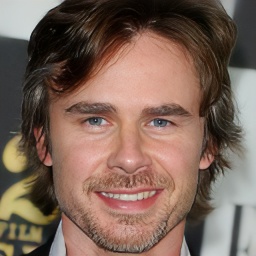}&   
     \includegraphics[width=0.132\linewidth]{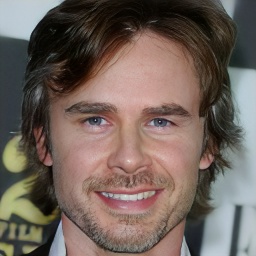} &        
     \includegraphics[width=0.132\linewidth]{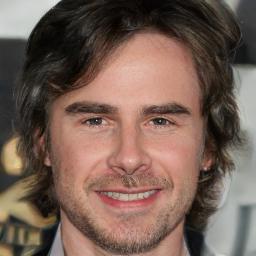} &
     \includegraphics[width=0.132\linewidth]{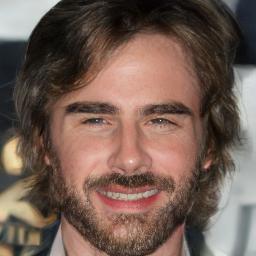}\\     
   
       \rotatebox{90}{ \hspace{6mm } + Age} &  
     \includegraphics[width=0.132\linewidth]{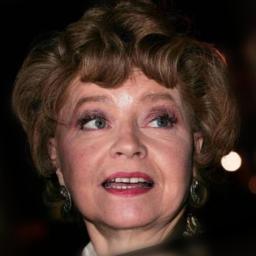}& 
     \includegraphics[width=0.132\linewidth]{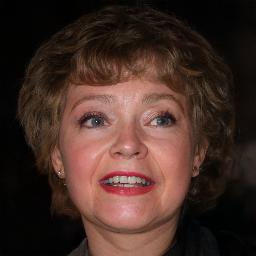}&   
     \includegraphics[width=0.132\linewidth]{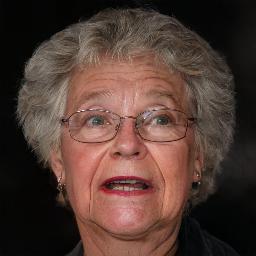} &   
     \includegraphics[width=0.132\linewidth]{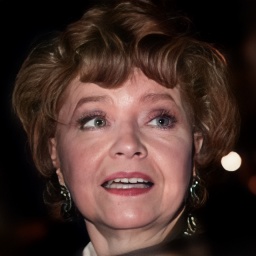}&   
     \includegraphics[width=0.132\linewidth]{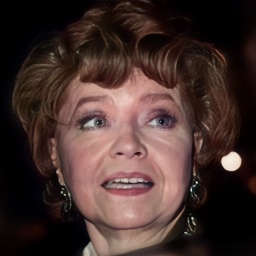} &         
     \includegraphics[width=0.132\linewidth]{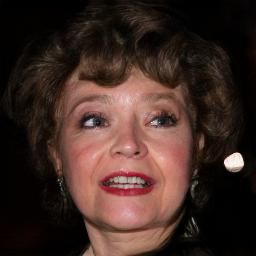} &   
     \includegraphics[width=0.132\linewidth]{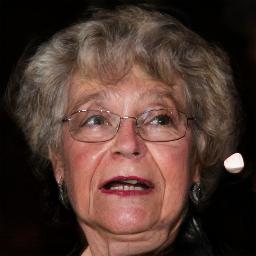} \\     
    & Source image & Rec by (a)  & Edit by (a)   & Rec by (b)  & Edit by (b)    & Rec by Ours & Edit by Ours  \\
    & Latent code  & Low-Rate & Low-Rate & Naive High-Rate & Naive High-Rate & Consultation  & Consultation  \\
    \end{tabular}
    \vspace{-2mm}
    \caption{Rate-Distortion-Edit trade-off. ``Rec'' and ``Edit'' represent the reconstruction and editing results, respectively. (a) is a typical low-rate framework for GAN inversion but suffers detail loss and distortion. (b) is a naive high-rate GAN inversion framework with nearly perfect reconstruction but suffers inferior interpretability and editability. The proposed method (Fig.~\ref{fig:overview}) combines both high details fidelity and compelling editing performance with a fast inference speed.}
    \vspace{-2mm}
    \label{fig:tradeoff}
\end{figure*}  

\section{Related Work} 
\subsection{GAN Inversion} 
Existing GAN inversion approaches can be categorized into optimization-based, encoder-based, and hybrid methods.  Optimization approaches can achieve high reconstruction quality but are slow for inference. \cite{zhu2016generative} used L-BFGS, and I2S ~\cite{abdal2019image2stylegan} adopted ADAM for solving the optimization. \cite{huh2020transforming} adopted   Covariance Matrix Adaptation   for gradient-free optimization.   Instead of per-image optimization, \cite{zhu2016generative} learned an encoder to project images. \cite{zhu2020domain}  proposed an in-domain method on real images.  pSp~\cite{richardson2020encoding} and GHFeat~\cite{xu2020generative} proposed to   embed latent codes in a hierarchical manner.  Further, e4e~\cite{tov2021designing} analyzed the trade-offs between reconstruction and editing ability. \cite{wei2021simple} improved the inversion efficiency by a shallow network with efficient heads. ReStyle~\cite{alaluf2021restyle} projected the latent codes with  iterative refinements. These methods are more efficient but fail to achieve high-fidelity reconstruction. Hybrid approaches make a compromise. \cite{zhu2016generative} initialize the optimization with the encoder output for acceleration. \cite{guan2020collaborative} designed a collaborative learning scheme for encoder and optimization iterator. \cite{roich2021pivotal}  fine-tuned StyleGAN parameters for each image after predicting an initial latent code, which takes a few minutes for an image.  Compared with previous methods, our method considerably improves the reconstruction quality of encoder models without inference-time optimization.

GAN inversion approaches can also be classified by the used latent space. $\mathbf{Z}$ space~\cite{karras2019style} is straightforward but suffers from feature entanglement.  $\mathbf{W}$~\cite{karras2019style} and $\mathbf{W}^+$~\cite{abdal2019image2stylegan,abdal2020image2stylegan++}  space in StyleGAN  are more disentangled, where $\mathbf{W}^+$ space  extends $\mathbf{W}$ space by using different ${W}$ across layers.  $\mathbf{S}$ space~\cite{wu2020stylespace} is  proposed  by transforming $\mathbf{W}^+$  through the affine layers.  $\mathbf{P}$ space~\cite{zhu2020improved}  inverts images to the last activation layer in the non-linear mapping network. Besides StyleGAN, some works~\cite{gu2020image} also adopts multi-scale latent codes for ProgressGAN~\cite{karras2018progressive}. Nevertheless, these latent spaces would inevitably lose details in reconstructed images due to limited bit-rate (Sec. \ref{3.1}). To perform a high-fidelity inversion, we propose a  distortion consultation branch to convey high-frequency image-specific information.

\subsection{Latent Space Editing} 
A number of supervised and unsupervised approaches explored GAN latent space for semantic directions under the vector arithmetic.  The supervised methods need off-the-shelf attribute classifiers or annotated images for specific attributes. InterfaceGAN~\cite{shen2020interpreting}   trained SVM to learn the boundary hyperplane for each binary attribute.  StyleFlow~\cite{abdal2020styleflow} learned reversible mapping by normalizing flow and off-the-shelf classifiers. Others~\cite{jahanian2019steerability,plumerault2020controlling}  explored  simple geometric transformation via self-supervised learning. Unsupervised approaches do not need pre-trained classifiers. GANspace~\cite{harkonen2020ganspace} performed PCA on early feature layers. Similarly, SeFa~\cite{shen2020interpreting} performed eigenvector decomposition of the affine layers. Some~\cite{voynov2020unsupervised,lu2020unsupervised,zhuang2021enjoy} found distinguishable directions based on mutual information. LatentCLR~\cite{yuksel2021latentclr} explored  directions  by contrastive learning.

\begin{figure*}[t]
    \centering
    \includegraphics[width=0.98\linewidth]{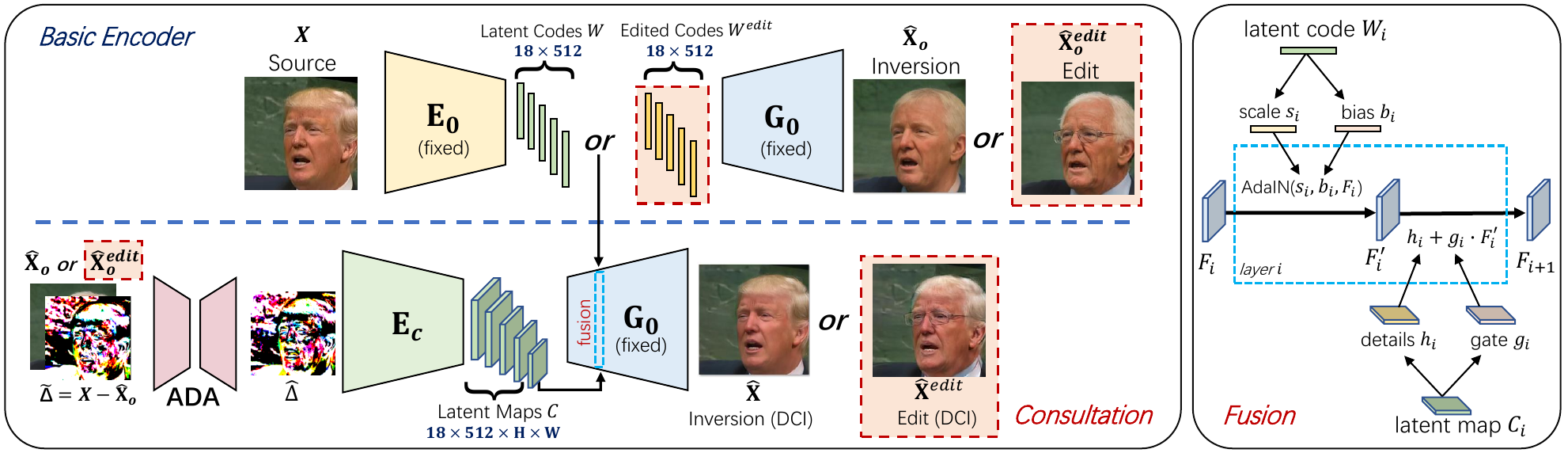}
    \caption{Overview of our high-fidelity image inversion and editing framework.  The basic encoder $E_0$ infers a low-rate latent code $W$ corresponding to a low-fidelity reconstruction image $\hat{X}_o$. The distortion map $\Tilde{\Delta}$ between $\hat{X}_o$ and the source image $X$ contains the lost high-frequency image-specific details to improve the reconstruction fidelity. The red dotted boxes indicate the editing behaviour with certain semantic direction $N^{edit}$, where $W^{edit} = W + \alpha N^{edit}$ corresponds to a low-fidelity editing image  $\hat{X}_o^{edit}$.  To achieve high-fidelity image editing, we propose the distortion consultation branch to facilitate the generation. In the distortion consultation, $\Tilde{\Delta}$  is first aligned with the low-fidelity edited image $\hat{X}_o^{edit}$ by ADA and then embedded to a high-rate latent map $C$ via the consultation encoder $E_c$. Latent code $W$ and latent map $C$ are combined via the consultation fusion (see details in the right part) across layers of $G_0$ to generate the final edited image $\hat{X}^{edit}$.}
    \label{fig:overview}
    \vspace{-2mm}
\end{figure*}

\section{Approach} 
\vspace{-1mm}
Given a source image $X$ and a well-trained generator $G_0$, GAN inversion infers the latent code $W$ via an encoder $E_0$, which is expected to faithfully reconstruct $X$. In this section, we first analyze the bottleneck of previous inversion methods and describe our proposed distortion consultation inversion strategy. To handle the features misalignment, we present the adaptive distortion alignment modules with a self-supervised training scheme. The whole framework is illustrated in Figure~\ref{fig:overview}.

\subsection{Overview} 
\label{3.1}
\noindent\textbf{Motivation.} Currently, GAN inversion frameworks lie in three categories, which are optimization-based, encoder-based and hybrid methods. Despite more accurate, optimization-based and hybrid approaches are time-consuming and thus intolerable in real-time applications. Existing encoder-based methods can be illustrated by Fig.~\ref{fig:tradeoff} (a), where the decoder is a frozen well-trained generator (e.g., StyleGAN) while the encoder learns a mapping from the source image to the latent codes. As observed in many existing works (e.g., results (a) in Fig. \ref{fig:tradeoff}), the encoder approaches fail to faithfully reconstruct the input images, and the inversion (and editing) results are of low fidelity in terms of details. Noted the fact that the latent codes in previous methods are of (relatively) low dimension, we conjecture that the low-rate latent codes are insufficient for high-fidelity reconstruction. This conjecture is also supported by the Rate-Distortion theory~\cite{shannon1959coding,cover1999elements,blau2019rethinking}, which will be reviewed in the \textbf{Supplement}.

To further analyze the effect of the latent rate in high-fidelity GAN inversion, we formulate the encoder-based GAN inversion as a problem of lossy data compression. In this formulation, \textit{rate} can be interpreted as the dimension of latent codes (e.g., $18\times512$), and \textit{distortion} indicates the reconstruction  quality (fidelity). A compelling inversion method is desired to produce high-fidelity images for both inversion and editing (\textit{low distortion}). Nevertheless, the current dimension of latent codes is much smaller than that of images (\textit{low rate}). This implies a contradiction with ~\cite{shannon1959coding,tishby2000information}, which shows the low-rate latent codes are insufficient for faithful reconstruction and some information is inevitably lost. Therefore, we are motivated to design a large-rate GAN inversion system.\\

\vspace{-3.5mm}
 \noindent\textbf{Challenge.} However, it is non-trivial to reduce the distortion by simply increasing the latent rate. A naive idea for faithful reconstruction is to adopt a higher-rate latent code like Fig.~\ref{fig:tradeoff} (b). This  Unet-like structure is adopted by some recent image restoration works~\cite{ChenPSFRGAN,wang2021gfpgan} that conveys  latent maps (e.g., $18\times512\times H\times W$) to decoder. Benefited from the higher bit rate, the restoration quality is gratifying (e.g., results (b) in Fig.~\ref{fig:tradeoff}). However, we cannot apply this structure in our case since the high-dimensional latent codes are difficult to interpret and manipulate for attribute editing (e.g., results (b) in Fig.~\ref{fig:tradeoff}). Similarly, prior work~\cite{tov2021designing} also observed tradeoffs between the reconstruction and editability brought by over-fitting. The high-rate latent code is easy to overfit on the reconstruction, thereby compromising the edit performance.   As the inversion is just an intermediate step to achieve the goal of editing, it is essential to balance the rate, reconstruction, and editing quality, which we call the \textit{Rate-Distortion-Edit} trade-offs (Fig.~\ref{fig:tradeoff}). To  this end, a delicate system design is needed.\\
 
\vspace{-3.3mm}
 \noindent\textbf{Design.} As analyzed above, with a (relatively) low-rate latent code, the GAN inversion system is subject to inevitable information loss. By analyzing the visual results of  previous GAN inversion approaches (Fig.~\ref{fig:teaser}, Fig.~\ref{fig:tradeoff}, Fig.~\ref{fig:editing-face}), we found that these reconstruction results can  successfully preserve frequent patterns and principle attributes of the source images. In contrast, the lost information is mostly the image-specific details such as background, make-up and illumination. This observation is  consistent with the Information Bottleneck theory~\cite{tishby2000information,tishby2015deep,shwartz2017opening}, which hypothesizes the deep models primarily learn common patterns in the dataset while forgetting infrequent details for reconstruction.  

 Considering the \textit{Rate-Distortion-Edit} trade-offs, now that we have prioritized the editability (with a low-rate latent code), the main concern is how to convey the lost information to improve the fidelity (lower the distortion) without compromising the edit performance. To this end, we propose a distortion consultation branch that only conveys image-specific details to enhance the reconstruction quality, which avoids the trivial solution of a simple overfitting. For editing, we still  perform vector arithmetic on the low-rate latent code for its high editability. By combining the best of both worlds, the proposed approach achieves a high fidelity in both reconstruction and editing (Fig.~\ref{fig:tradeoff}).

\subsection{Distortion Consultation Inversion (DCI)} 
\noindent\textbf{Basic Encoder.} With a basic encoder $E_0$, we can obtain a low-rate latent code $W =  E_0(X)$ and initial inversion image  $\hat{X}_o = G_0 (W)$. In this case, the  generator $G_0$ takes $W$ as the input  in each layer to obtain the feature map:
 \begin{equation}
\begin{aligned}
F_{i+1} &=   AdaIN(F_i, f ^{s}_i(W_i), f ^{b}_i(W_i)),
\end{aligned}
\end{equation}
where $f ^{s}_i(W_i), f ^{b}_i(W_i)$ are affine layers for scale and bias in $AdaIN$ \cite{Huang2017}. $\hat{X}_o$ is  low-fidelity due to the information loss of low-rate  latent codes, and the subscript $o$ denotes an (unsatisfactory) observation of source image $X$.

\noindent\textbf{Consultation Encoder.} To enhance $E_0$ with higher fidelity, we propose a distortion consultation branch  to convey the lost image-specific details.  We refer it to  \textbf{\textit{Consultation}}, since the network explicitly consults the image-specific information as a reference for generation. Specifically,  we see  the distortion map $\Tilde{\Delta} = X - \hat{X}_o $  between source $X$ and initial reconstruction $\hat{X}_o$  as the lost details~\cite{Wang_2021_ICCV}.  The distortion map is projected to a high-rate latent map $C=E_c(\Tilde{\Delta})$ via the consultation encoder $E_c$. Compared with prior methods relying on $W$ only,  $G_0$ additionally consults $C$ for lost details to achieve high-fidelity reconstruction as $\hat{X}=G_0(W, C)$. \\

\label{3.2}
\vspace{-2.5mm}
\noindent\textbf{Consultation Fusion.} To combine the consultation branch with the basic encoder for image generation, we adopt a layer-wise consultation fusion for latent codes $W$ and latent maps $C$, as shown in Fig.~\ref{fig:overview}. As artifacts and inaccurate details introduced by $W$ can degrade the generation quality, we design a gated fusion scheme to adaptively filter out undesired  features. In layer $i$ of $G_0$,  $C_i$ is embedded to a gate map $g_i$ and a  high-frequency details map $h_i$:
\begin{equation}
\begin{aligned}
g_{i} &= f ^{gate}_i (C_i), ~  h_{i} = f ^{hf}_i (C_i),
\end{aligned}
\end{equation}
where mapping functions $f^{gate}$ and $f^{hf}$ are convolution layers. $h_i$ contains the image-specific details, and facilitates the low-fidelity features obtained from $W_i$ (Eq. (2)) to produce high-fidelity feature maps $F_{i+1}$   in StyleGAN\footnote{For StyleGAN2~\cite{karras2020analyzing}, the fusion layer would be $F_{i+1}  =  g_i \cdot ModulatedConv(F_i, f ^{s}_i(W_i), f ^{b}_i(W_i)) + h_i$.}:
 \begin{equation}
\begin{aligned}
F_{i+1} &= g_i \cdot AdaIN(F_i, f ^{s}_i(W_i), f ^{b}_i(W_i)) + h_i.
\end{aligned}
\end{equation}
To avoid overfitting on the inversion result, we only perform the consultation fusion in early layers of $G_0$.

\subsection{Adaptive Distortion Alignment (ADA)} 
For attribute editing, the low-rate latent code $W$ would be moved along certain semantic direction $N^{edit}$ as  $W^{edit} = W + \alpha N^{edit}$~\cite{shen2020interpreting}. The initial edited image by the basis encoder is denoted as $\hat{X}_o^{edit} = G_0(W^{edit})$, which suffers details distortion. So far, we have improved the fidelity of the inversion image $\hat{X}_o$ with the proposed DCI, where  the distortion map  $\Tilde{\Delta} = X - \hat{X}_o $ is calculated for $\hat{X}_o$. However,  $\hat{X}_o^{edit}$ would be deformed from  $\hat{X}_o$ when editing attributes such as age, pose and expression. This means  the observed $\Tilde{\Delta}$ may not align with the edited image $\hat{X}_o^{edit}$. Applying DCI directly to $\hat{X}_o^{edit}$ leads to obvious artifacts by consulting misaligned details $\Tilde{\Delta}$ (see Sec. \ref{4.3}). To advance DCI from inversion to editing, the observed distortion map $\Tilde{\Delta}$ is supposed to be adaptively aligned with the edited image $\hat{X}_o^{edit}$. We thus propose the ADA module, which is an encoder-decoder-like structure for distortion alignment.

Considering a misaligned pair  of  \{$I$,  $\Tilde{\Delta}$\} where $I$ is  $\hat{X}_o$ for inversion and  $\hat{X}_o^{edit}$ for editing, ADA is to align the distortion map $\Tilde{\Delta}$ with a target image $I$. For inversion, ADA is ideally an identity mapping. For editing, the distortion map is desired to be adaptively transformed  as $\hat{\Delta}^{edit} = ADA(\hat{X}^{edit}_o, \Tilde{\Delta})$ that aligns with the initial editing result $\hat{X}_o^{edit}$. With $C^{edit} = E_c (\hat{\Delta}^{edit})$ as a reference, $\hat{X}^{edit}=G_0(W^{edit}, C^{edit})$ can preserve more  details.\\ 

\vspace{-2.5mm}
\noindent\textbf{Self-supervised Training.} To alleviate the entanglement between distortion alignment and distortion consultation, involving intermediate supervisions on  ADA outputs in preferred. To this end, we need numerous misaligned pairs of  \{$I$,  $\Tilde{\Delta}$\} and their ground-truth aligned maps $\Delta$ for training, but the data collection is labor-intensive. To conduct a self-supervised training, we take  $X$ as the source image, and the low-fidelity inversion $\hat{X}_o$ as the target image $I$ for alignment, and the ground-truth aligned distortion is thus $\Delta = X - \hat{X}_o$.  During the training, we augment $\Delta$ with random perspective transformation to simulate misaligned distortion maps $\Tilde{\Delta}$. We empirically observe these simulated pairs work well on training and expect a better simulation scheme in future works. The ADA module is encouraged to produce the aligned distortion $\hat{\Delta} = ADA(\hat{X}_o, \Tilde{\Delta})$ that approximates  $\Delta$. The alignment loss is defined as: $ \mathcal{L}_{align} = \|\hat{\Delta}- \Delta \|_{1} .$ See the \textbf{Supplement} for more details.

\subsection{Losses} 
During the training, the generator $G_0$ and basic encoder $E_0$ are frozen.  For faithful reconstruction, we calculate  $L_2$ loss and LPIPS~\cite{zhang2018perceptual}  between  $\hat{X}$ and $X$. We also calculate the identity loss  $\mathcal{L}_{\mathrm{id}} = 1- \langle  F(X), F( \hat{X}) \rangle$, where $F$ is pre-trained ArcFace~\cite{deng2019arcface} or a ResNet-50 model for different domains~\cite{tov2021designing}. The reconstruction loss is
\begin{equation}
\mathcal{L}_{rec} = \mathcal{L}_{2} +\lambda_{per} \mathcal{L}_{LPIPS}  +\lambda_{id} \mathcal{L}_{id}  .
\end{equation}
We also impose adversarial loss to improve image quality: 
\begin{align}
\mathcal{L}_{D} &=   {\mathbb{E}}[\log D(\hat{X})]  +  {\mathbb{E}}[\log (1 - D(X) )],\\
\mathcal{L}_{adv} &= - {\mathbb{E}}[\log (D(\hat{X}))],
\end{align}
where $D$ is initialized with the well-trained discriminator.\\
In summary, the overall loss is a weighted summation of $\mathcal{L}_{rec}$, $\mathcal{L}_{adv}$, and $\mathcal{L}_{align}$.
Note that the training process only involves  the inversion images, and no editing direction $N^{edit}$ is needed. After training, the model can generalize to diverse attribute editing explored by different methods.

\begin{figure*}[!ht]
    \centering 
    \small
    \begin{tabular}{@{}c@{\hspace{0.2mm}}c@{\hspace{1.5mm}}c@{\hspace{0.2mm}}c@{\hspace{0.2mm}}c@{\hspace{0.2mm}}c@{\hspace{ 1.5mm}}c@{\hspace{0.2mm}}c@{\hspace{0.2mm}}c@{\hspace{0.2mm}}c@{}}
    \rotatebox{90}{\hspace{4mm} + Age }   &
     \includegraphics[width=0.104\linewidth]{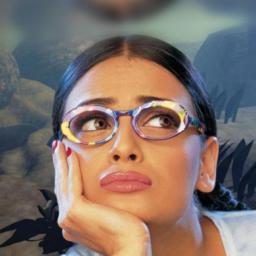}       &   
     \includegraphics[width=0.104\linewidth]{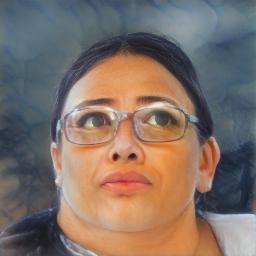}   &   
     \includegraphics[width=0.104\linewidth]{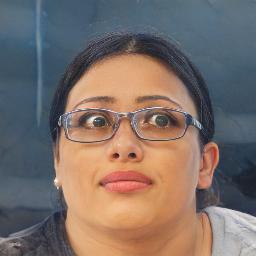}       &    
     \includegraphics[width=0.104\linewidth]{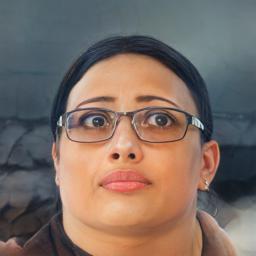}      &  
     \includegraphics[width=0.104\linewidth]{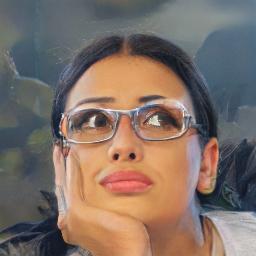}       &     
     \includegraphics[width=0.104\linewidth]{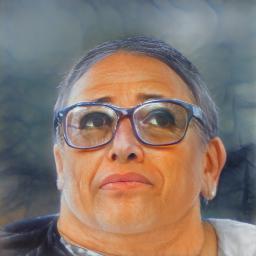}     &   
     \includegraphics[width=0.104\linewidth]{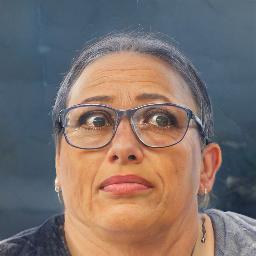}       &        
     \includegraphics[width=0.104\linewidth]{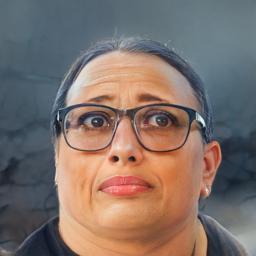}     &   
     \includegraphics[width=0.104\linewidth]{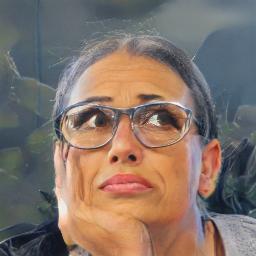}      \\

    \rotatebox{90}{\hspace{4mm} - Age}   &
     \includegraphics[width=0.104\linewidth]{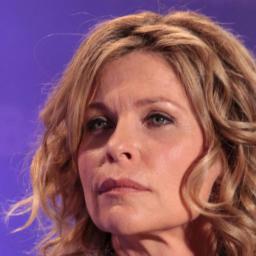}       &   
     \includegraphics[width=0.104\linewidth]{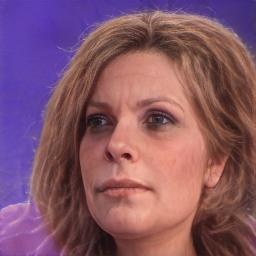}    &   
     \includegraphics[width=0.104\linewidth]{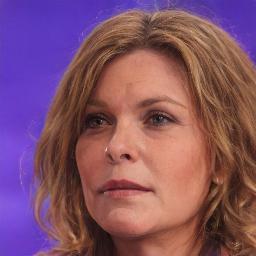}       &   
     \includegraphics[width=0.104\linewidth]{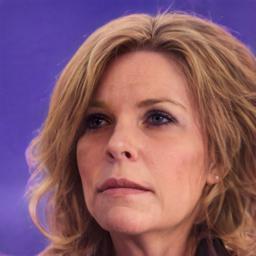}     &   
     \includegraphics[width=0.104\linewidth]{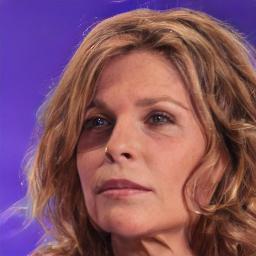}      & 
     \includegraphics[width=0.104\linewidth]{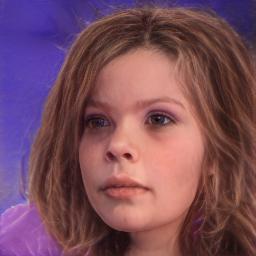}     &   
     \includegraphics[width=0.104\linewidth]{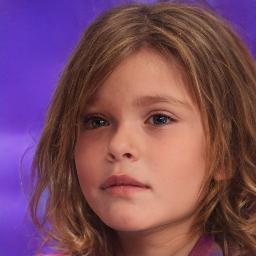}      &        
     \includegraphics[width=0.104\linewidth]{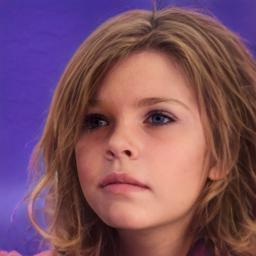}     &   
     \includegraphics[width=0.104\linewidth]{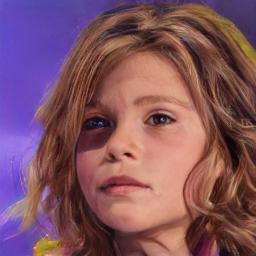}     \\
     
    \rotatebox{90}{\hspace{3mm}  + Smile }   &
     \includegraphics[width=0.104\linewidth]{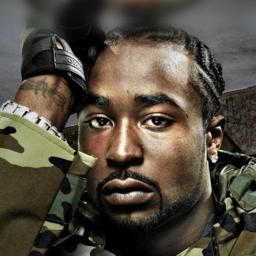}  &   
     \includegraphics[width=0.104\linewidth]{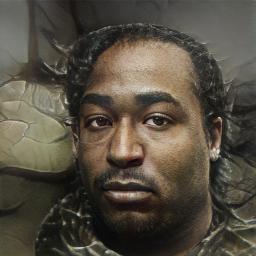}   &   
     \includegraphics[width=0.104\linewidth]{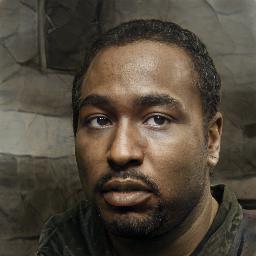}     &
     \includegraphics[width=0.104\linewidth]{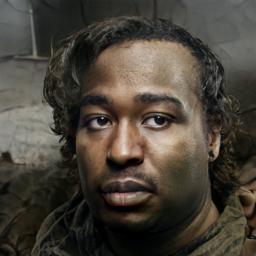}    &
     \includegraphics[width=0.104\linewidth]{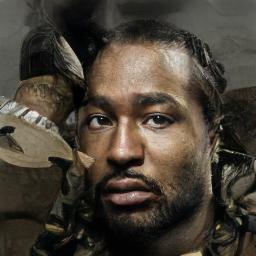}    &   
     \includegraphics[width=0.104\linewidth]{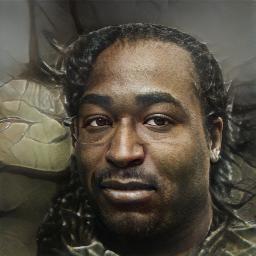}   &   
     \includegraphics[width=0.104\linewidth]{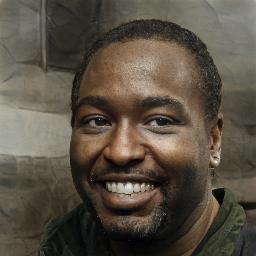}   &        
     \includegraphics[width=0.104\linewidth]{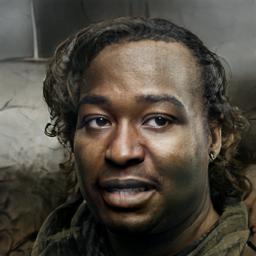}    &   
     \includegraphics[width=0.104\linewidth]{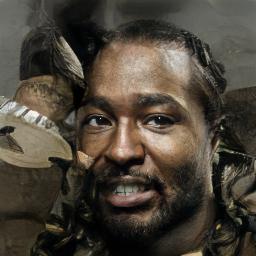}  \\

    \rotatebox{90}{\hspace{3mm}  - Smile }   &
     \includegraphics[width=0.104\linewidth]{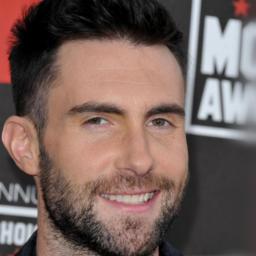}    &   
     \includegraphics[width=0.104\linewidth]{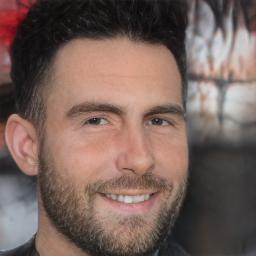}   &    
     \includegraphics[width=0.104\linewidth]{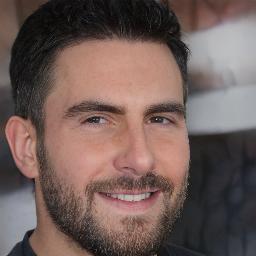}    &  
     \includegraphics[width=0.104\linewidth]{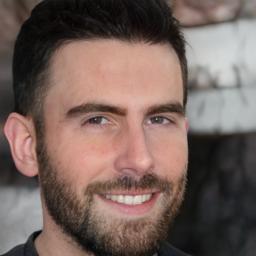}     &  
     \includegraphics[width=0.104\linewidth]{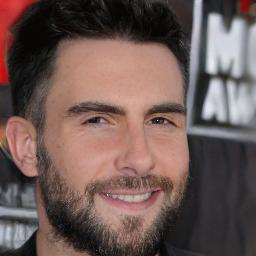}   &    
     \includegraphics[width=0.104\linewidth]{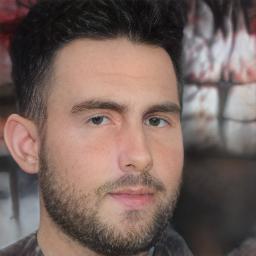}      &   
     \includegraphics[width=0.104\linewidth]{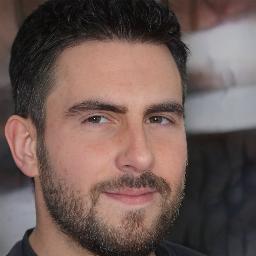}   &        
     \includegraphics[width=0.104\linewidth]{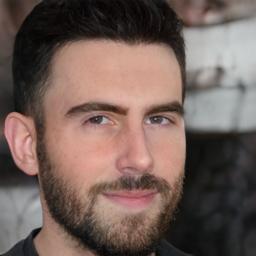}     &   
     \includegraphics[width=0.104\linewidth]{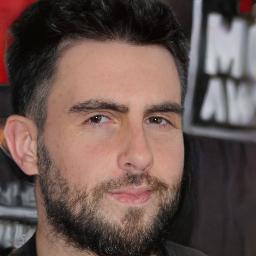}  \\   
     
     \rotatebox{90}{\hspace{1.2mm}  + Lip Stick }   &
     \includegraphics[width=0.104\linewidth]{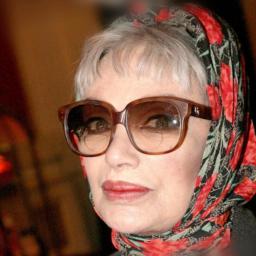}  &   
     \includegraphics[width=0.104\linewidth]{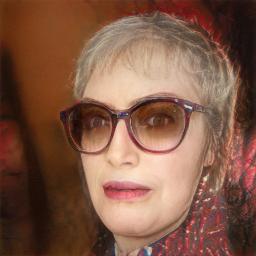}     &
     \includegraphics[width=0.104\linewidth]{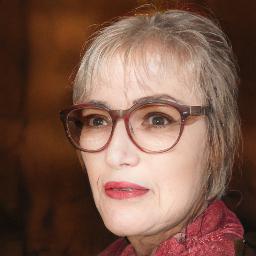}    &   
      \includegraphics[width=0.104\linewidth]{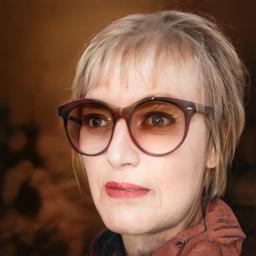}     & 
     \includegraphics[width=0.104\linewidth]{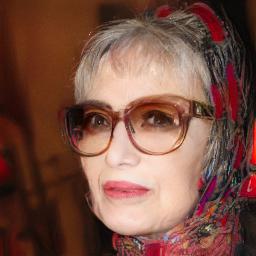}      &
     \includegraphics[width=0.104\linewidth]{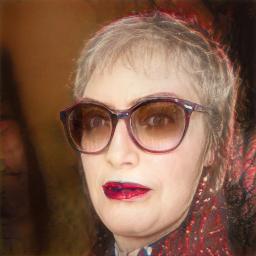}      &   
     \includegraphics[width=0.104\linewidth]{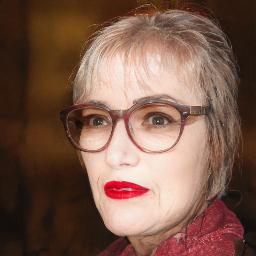}  &        
     \includegraphics[width=0.104\linewidth]{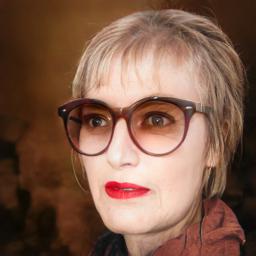}      &   
     \includegraphics[width=0.104\linewidth]{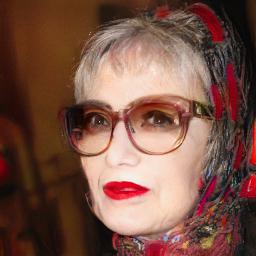}\\

    \rotatebox{90}{\hspace{4mm}   + Pose }   &
     \includegraphics[width=0.104\linewidth]{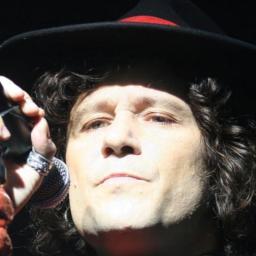} &   
     \includegraphics[width=0.104\linewidth]{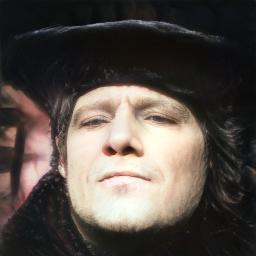}      &     
    \includegraphics[width=0.104\linewidth]{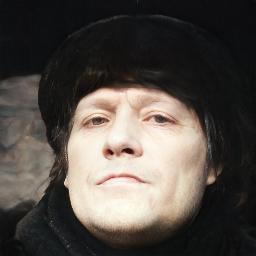}   & 
     \includegraphics[width=0.104\linewidth]{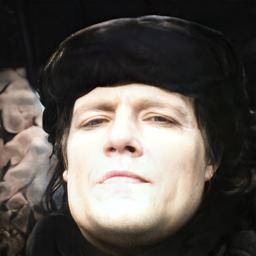}     & 
     \includegraphics[width=0.104\linewidth]{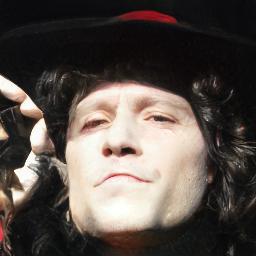}    & 
     \includegraphics[width=0.104\linewidth]{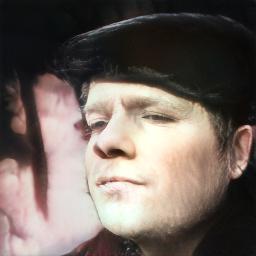}     &   
     \includegraphics[width=0.104\linewidth]{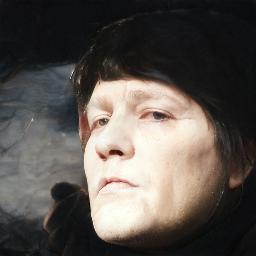} &        
     \includegraphics[width=0.104\linewidth]{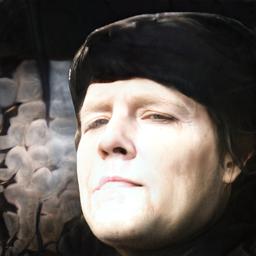}     &   
     \includegraphics[width=0.104\linewidth]{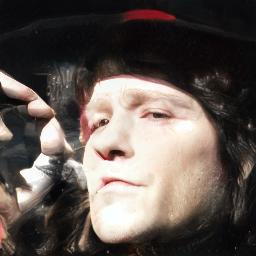}\\ 
     
    \rotatebox{90}{\hspace{4mm}  - Pose }   &
     \includegraphics[width=0.104\linewidth]{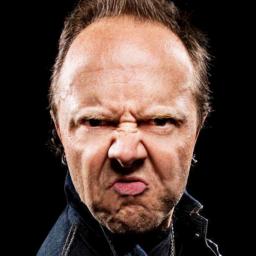} &   
     \includegraphics[width=0.104\linewidth]{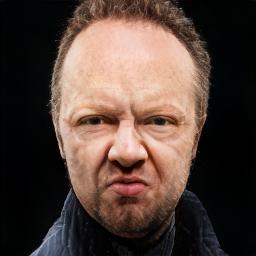}     &       
    \includegraphics[width=0.104\linewidth]{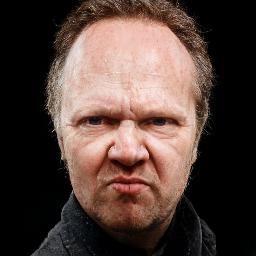}  & 
     \includegraphics[width=0.104\linewidth]{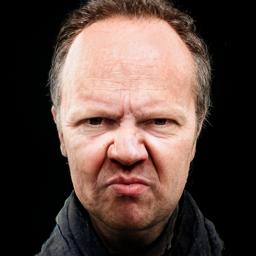}      &
     \includegraphics[width=0.104\linewidth]{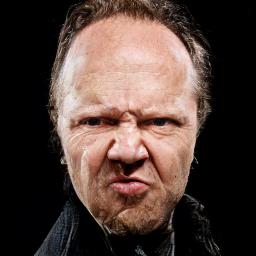}    &  
     \includegraphics[width=0.104\linewidth]{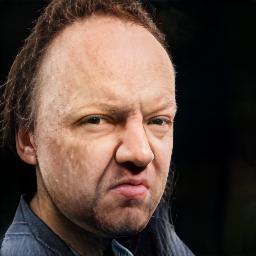}     &   
     \includegraphics[width=0.104\linewidth]{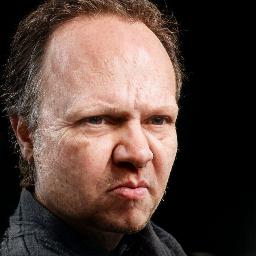}  &        
     \includegraphics[width=0.104\linewidth]{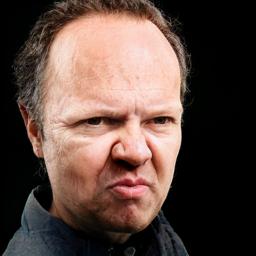}   &   
     \includegraphics[width=0.104\linewidth]{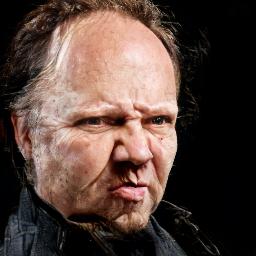}\\  
     
    \rotatebox{90}{\hspace{1.4mm}  Close Eyes }   &
     \includegraphics[width=0.104\linewidth]{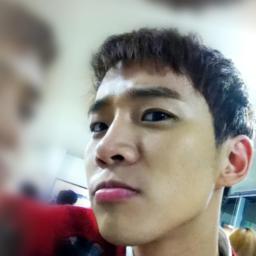} &   
     \includegraphics[width=0.104\linewidth]{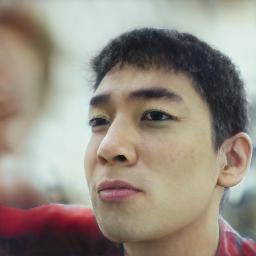}     &  
     \includegraphics[width=0.104\linewidth]{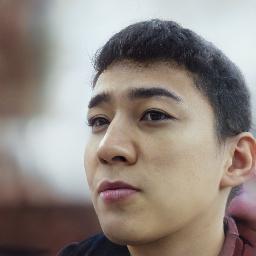}  &    
    \includegraphics[width=0.104\linewidth]{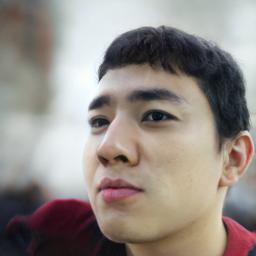}    & 
    \includegraphics[width=0.104\linewidth]{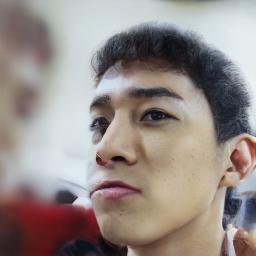}    & 
     \includegraphics[width=0.104\linewidth]{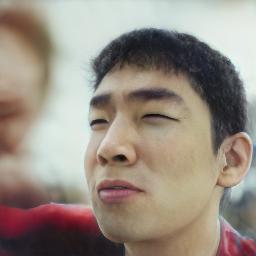}      &   
     \includegraphics[width=0.104\linewidth]{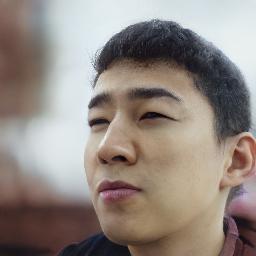} &        
     \includegraphics[width=0.104\linewidth]{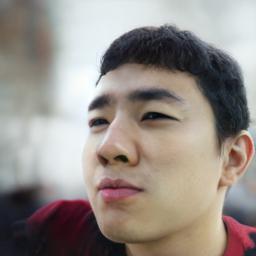}     &   
     \includegraphics[width=0.104\linewidth]{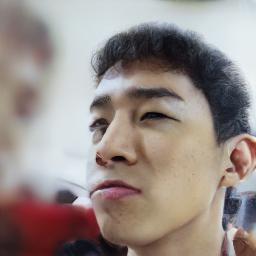} \\      

    \rotatebox{90}{\hspace{3mm}  Color }   &
     \includegraphics[width=0.104\linewidth]{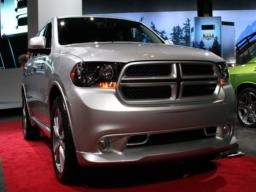}  &   
     \includegraphics[width=0.104\linewidth]{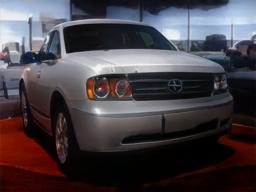} &  
     \includegraphics[width=0.104\linewidth]{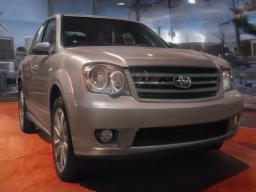}  & 
     \includegraphics[width=0.104\linewidth]{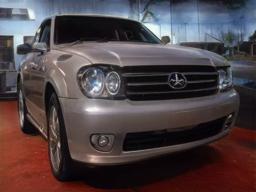}  &  
     \includegraphics[width=0.104\linewidth]{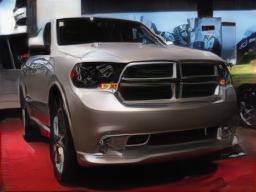}  &
     \includegraphics[width=0.104\linewidth]{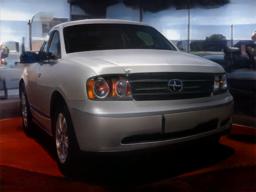}  &  
     \includegraphics[width=0.104\linewidth]{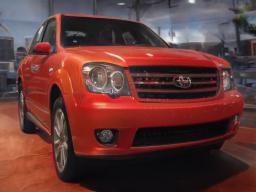} &     
     \includegraphics[width=0.104\linewidth]{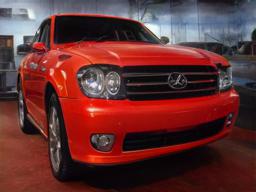}   &   
     \includegraphics[width=0.104\linewidth]{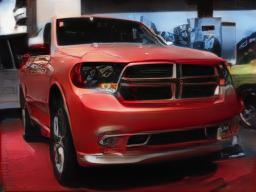} \\

    \rotatebox{90}{\hspace{3mm}  Color }   &
     \includegraphics[width=0.104\linewidth]{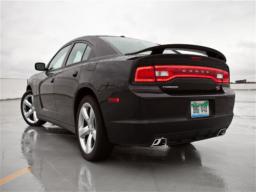}  &   
     \includegraphics[width=0.104\linewidth]{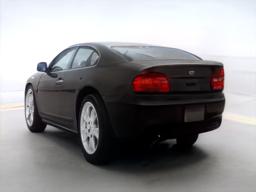}  &  
     \includegraphics[width=0.104\linewidth]{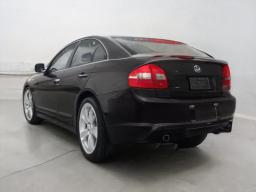} & 
      \includegraphics[width=0.104\linewidth]{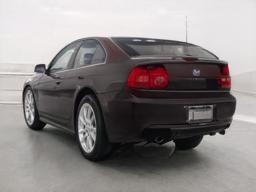} &   
     \includegraphics[width=0.104\linewidth]{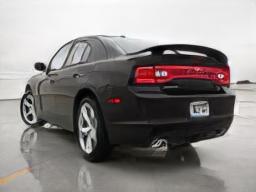}  &  
     \includegraphics[width=0.104\linewidth]{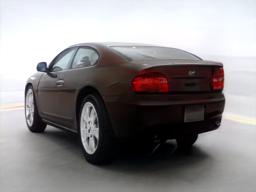}  &  
     \includegraphics[width=0.104\linewidth]{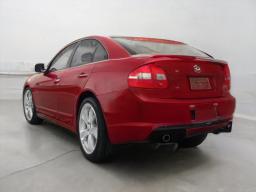}  &     
     \includegraphics[width=0.104\linewidth]{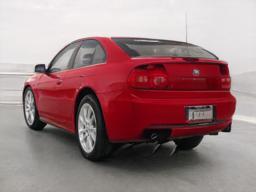}   &   
     \includegraphics[width=0.104\linewidth]{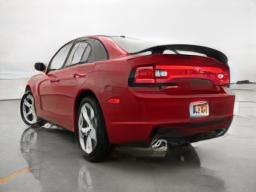} \\      

    \rotatebox{90}{\hspace{1.7mm} + Grass }   &
     \includegraphics[width=0.104\linewidth]{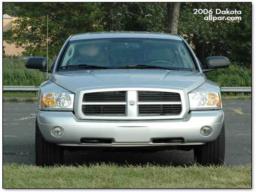} &   
     \includegraphics[width=0.104\linewidth]{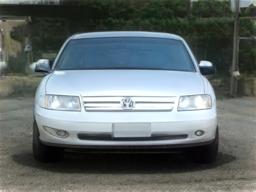} &  
    \includegraphics[width=0.104\linewidth]{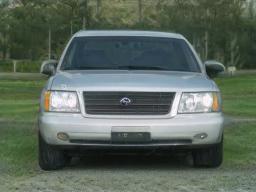} &  
     \includegraphics[width=0.104\linewidth]{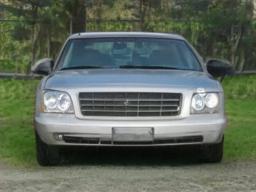} &   
     \includegraphics[width=0.104\linewidth]{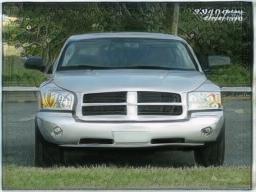}  & 
     \includegraphics[width=0.104\linewidth]{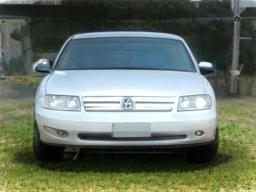}  &  
     \includegraphics[width=0.104\linewidth]{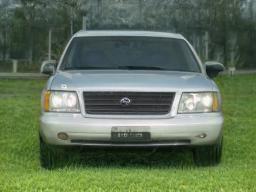}  &     
     \includegraphics[width=0.104\linewidth]{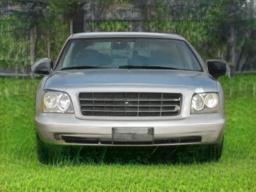}   &   
     \includegraphics[width=0.104\linewidth]{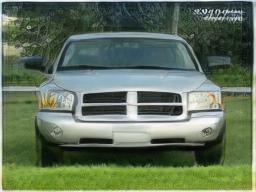} \\
     
    \rotatebox{90}{\hspace{1.7mm} + Grass }   &
     \includegraphics[width=0.104\linewidth]{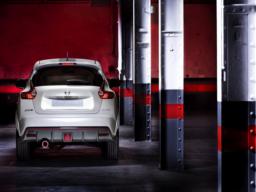} &   
     \includegraphics[width=0.104\linewidth]{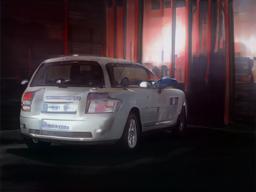} &  
      \includegraphics[width=0.104\linewidth]{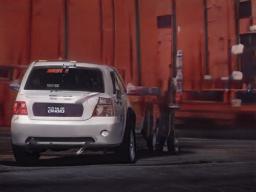} & 
     \includegraphics[width=0.104\linewidth]{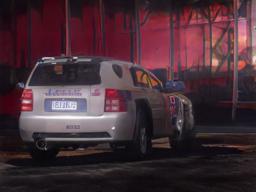} &  
      \includegraphics[width=0.104\linewidth]{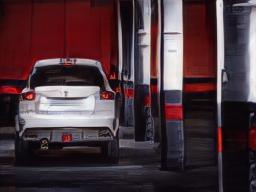}  &  
     \includegraphics[width=0.104\linewidth]{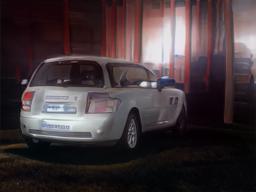}  &  
     \includegraphics[width=0.104\linewidth]{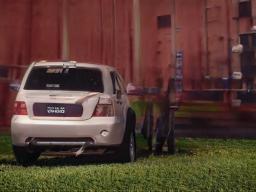}  &     
     \includegraphics[width=0.104\linewidth]{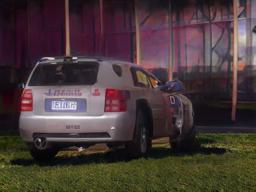}   &   
     \includegraphics[width=0.104\linewidth]{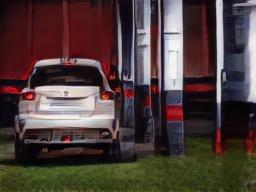} \\      
     
    & Input&pSp (Rec)&  e4e (Rec)& Restyle (Rec) &Ours (Rec) & pSp (Edit) & e4e (Edit)&    Restyle (Edit)  &Ours (Edit)
    \end{tabular}
   \caption{Visual comparisons on Face  inversion and editing. More results are shown in \textbf{Appendix}.}
    \label{fig:editing-face}
\end{figure*} 

\vspace{-1mm}
\section{Experiments} 
\vspace{-1mm}
\subsection{Settings} 
\vspace{-1mm}
\noindent\textbf{Datasets.} For the human face domain, we use the FFHQ~\cite{karras2019style} dataset for training and the CelebA-HQ~\cite{karras2018progressive} dataset for evaluation. For the car domain, we use Stanford Cars~\cite{Krause_3DRR2013} for training and evaluation. For attribute editing, we adopt InterfaceGAN~\cite{shen2020interpreting} for  face images  and GANSpace~\cite{harkonen2020ganspace} for  car images.\\
\textbf{Implementation details. } See the \textbf{Appendix}.

\begin{table}[t]
\centering
\small
\renewcommand{\arraystretch}{1.1}
\begin{tabular}{@{ }l@{\hspace{3mm}}c@{\hspace{3mm}}c@{\hspace{3mm}}c@{\hspace{3mm}}c@{}}
\hline Method &  MAE  $\downarrow$ &  SSIM $\uparrow$ &   LPIPS $\downarrow$ &  Time $\downarrow$ \\
\hline
I2S~\cite{abdal2019image2stylegan} & .0636$\pm$.0010 & .872$\pm$.005 & .134$\pm$.006 & 156s \\
PTI~\cite{roich2021pivotal} & .0622$\pm$.0004 & .877$\pm$.003 & .132$\pm$.003 & 283s \\
\hline
pSp \cite{richardson2020encoding} &  .0789$\pm$.0006  & .793$\pm$.006 &  .169$\pm$.002 & 0.11s \\
Restyle$_{pSp}$ \cite{alaluf2021restyle} & .0729$\pm$.0005   &.823$\pm$.004   & .145$\pm$.002 & 0.46s \\
e4e \cite{tov2021designing}& .0919$\pm$.0008  & .742$\pm$.007 & .221$\pm$.003 &  0.11s \\
Restyle$_{e4e}$ \cite{alaluf2021restyle}& .0887$\pm$.0008  &.758$\pm$.007  & .202$\pm$.003 & 0.46s \\
\hline
Ours$_{e4e}$ & \textbf{.0617$\pm$.0004}  & \textbf{.877$\pm$.002}  & \textbf{.127$\pm$.001} & 0.24s \\
\hline
\end{tabular}
\vspace{-2mm}
\caption{\small{Quantitative comparison for inversion quality on faces.}}
\label{tab:metric}
\vspace{-5mm}
\end{table}

\subsection{Evaluation} 
\subsubsection{Quantitative Evaluation} 
\vspace{-2mm}
We compare our method (with e4e as the basic encoder) with state-of-the-art encoder-based GAN inversion approaches, pSp~\cite{richardson2020encoding}, e4e~\cite{tov2021designing} and Restyle~\cite{alaluf2021restyle} (with pSp and e4e as backbones, respectively).  We report quantitative comparisons of the inversion performance in Table~\ref{tab:metric}. The metrics are calculated on the first 1,500 images from CelebA-HQ. We also compare the proposed method with two optimization-based approaches~\cite{abdal2019image2stylegan,roich2021pivotal}. Our approach substantially outperforms encoder-based baselines in terms of reconstruction quality and is considerably faster than optimization-based methods when inference. 
\vspace{-2mm}
\subsubsection{Qualitative Evaluation} 
\begin{figure*}[t]
    \centering 
    \small
    \begin{tabular}{@{}c@{\hspace{2mm}}c@{}}    
   \rotatebox{90}{\hspace{3mm}  + Pose }  & \includegraphics[width= 0.95\linewidth]{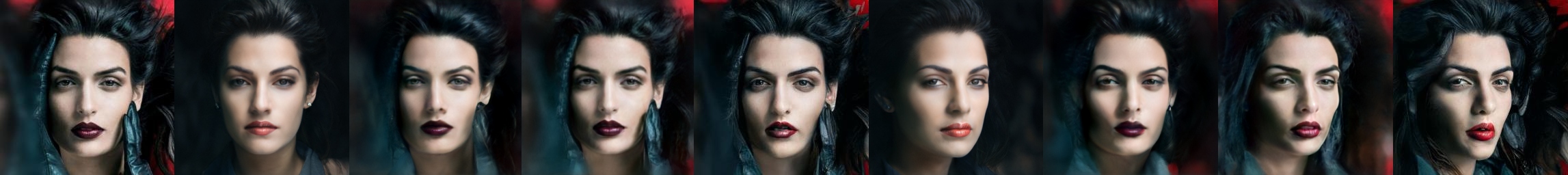}\\
   \rotatebox{90}{\hspace{3mm}  + Age }  & \includegraphics[width= 0.95\linewidth]{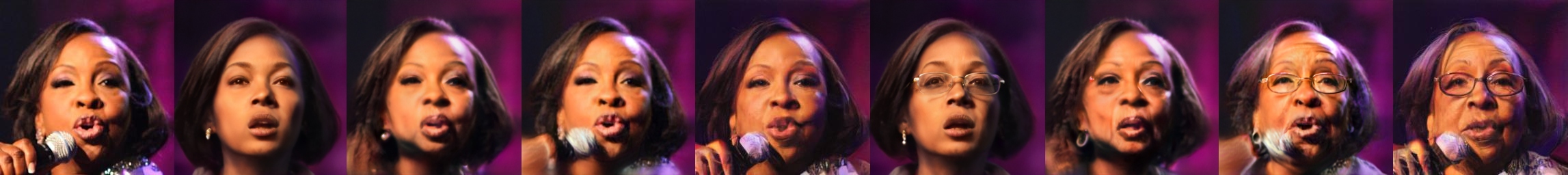}\\   
    \end{tabular}
    \begin{tabular}{@{}c@{\hspace{1.5mm}}c@{\hspace{0.2mm}}c@{\hspace{0.2mm}}c@{\hspace{0.2mm}}c@{\hspace{ 1.5mm}}c@{\hspace{0.2mm}}c@{\hspace{0.2mm}}c@{\hspace{0.2mm}}c@{}}
    \hspace{6mm} Input& \hspace{4mm} $\mathbb{W}$~\cite{karras2020analyzing} (Rec)& \hspace{1mm} $\mathbb{W}^+$~\cite{abdal2019image2stylegan} (Rec)& \hspace{1mm}PTI~\cite{roich2021pivotal} (Rec) &\hspace{1mm} Ours (Rec) & \hspace{3mm} $\mathbb{W}$ (Edit) &\hspace{5mm}  $\mathbb{W}^+$ (Edit)&    \hspace{4mm} PTI (Edit)  & \hspace{4mm} Ours (Edit)
    \end{tabular}
    \vspace{-2mm}
    \caption{Visual comparison with optimization-based methods. Our method is much faster than these baselines.}
    \vspace{-4mm}
    \label{fig:optimization}
\end{figure*}

\vspace{-2mm}
\noindent\textbf{Encoder baselines.} We show visual results of both inversion and editing in Fig.~\ref{fig:editing-face}. Compared with previous approaches, our method is robust to images with occlusion and extreme viewpoints. For example, the first row in Fig.~\ref{fig:editing-face} gives a face image occluded by the hand, and the last row demonstrates a car image with an out-of-range viewpoint. Existing methods fail to reconstruct these challenging images faithfully. They generate distorted results and suffer artifacts for both inversion and editing. In contrast, with the proposed distortion consultation scheme, our method is more robust with high-fidelity results. Besides the robustness improvement, our approach also successfully preserves more details in backgrounds (4th row), shadow (2nd row), reflect (10th row), accessory (5th row), expressions (7th and 8th rows), and appearance (9th and 11th rows). \\
\noindent\textbf{Optimization baselines.} We also compare our method with optimization-based methods~\cite{karras2020analyzing,abdal2019image2stylegan,roich2021pivotal} in Fig.~\ref{fig:optimization}. Note that PTI~\cite{roich2021pivotal} optimizes both latent codes and StyleGAN parameters, but we still report their results for better comparison. With  $\sim1000\times$ faster inference, our method achieves a comparable or even better reconstruction quality. Also, the editing results produced by the proposed scheme successfully preserve the image-specific details of source images without compromising the edit performance. 
\begin{table}[t]
\centering
\small
\renewcommand{\arraystretch}{1.1}
\begin{tabular}{@{}l@{\hspace{6mm}}c@{\hspace{4mm}}c@{\hspace{4mm}}c@{}}
\hline  &  Ours $>$ pSp& Ours $>$ e4e &  Ours $>$ Restyle \\
\hline
 Preference Rate& 81.2\% &  84.4\% &  79.7\% \\
\hline
\end{tabular}
\vspace{-1mm}
\caption{\small{The results of the user study. The reported value indicates the preference rate of Ours against a baseline.}}
\label{tab:user}
\vspace{-5mm}
\end{table}

\vspace{-3mm}
\subsubsection{User Study} 
\vspace{-2mm}
To perceptually evaluate the editing performance, we conduct a  user study in Table~\ref{tab:user}. We select the first 50 images from CelebA-HQ and perform editing on extensive attributes. We collect 1,500 votes from 30 participants. Each participant is given a triple of images (source, our editing, baseline editing) at once and asked to choose the higher-fidelity one with proper editing. The user study shows our method outperforms baselines by a large margin.

\begin{figure}[t]
    \centering 
    \small
    \begin{tabular}{@{}c@{\hspace{0.4mm}}c@{\hspace{0.2mm}}c@{\hspace{0.2mm}}c@{\hspace{0.2mm}}c@{}}
     \rotatebox{90}{\qquad Input}   &
     \includegraphics[width=0.205\linewidth]{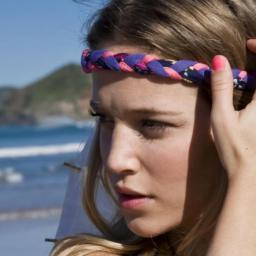}  &   
     \includegraphics[width=0.205\linewidth]{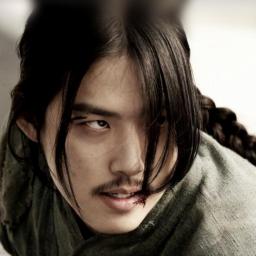}  &   
     \includegraphics[width=0.272\linewidth]{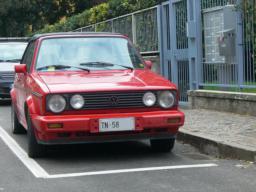}  &   
     \includegraphics[width=0.272\linewidth]{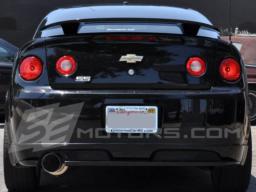}  \\     
     
     \rotatebox{90}{\quad w/o DCI }   &
     \includegraphics[width=0.205\linewidth]{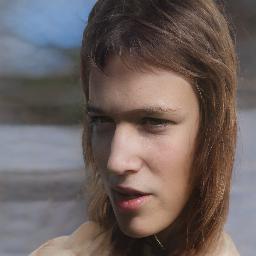} &       \includegraphics[width=0.205\linewidth]{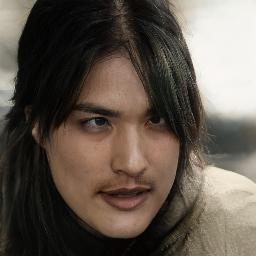} &  
     \includegraphics[width=0.272\linewidth]{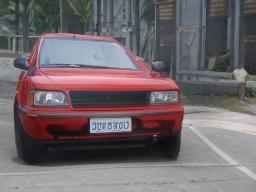} &      
     \includegraphics[width=0.272\linewidth]{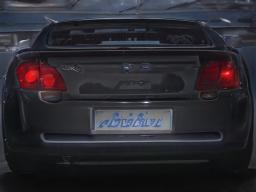} \\       
     
     \rotatebox{90}{\quad w/ DCI }   &
     \includegraphics[width=0.205\linewidth]{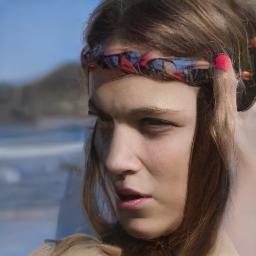}  &
     \includegraphics[width=0.205\linewidth]{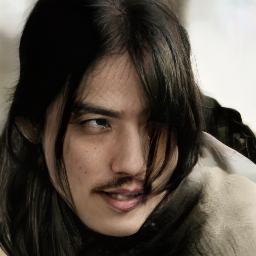}  &
     \includegraphics[width=0.272\linewidth]{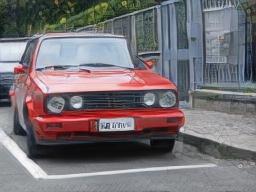}  &
     \includegraphics[width=0.272\linewidth]{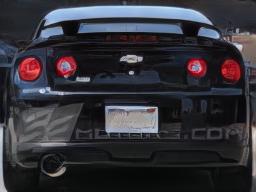}   \\
    \end{tabular}
    \vspace{-3mm}
   \caption{Effects of distortion consultation inversion (DCI). }
    \label{fig:inversion}
    \vspace{-2mm}
\end{figure}

\begin{figure}[t]
    \centering 
    \small
    \begin{tabular}{@{}c@{\hspace{0.4mm}}c@{\hspace{0.4mm}}c@{\hspace{0.4mm}}c@{\hspace{0.4mm}}c@{}}
     \includegraphics[width=0.195\linewidth]{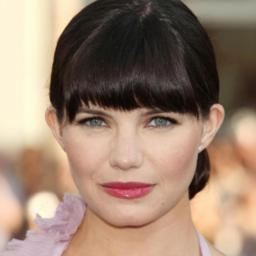} &   
    \includegraphics[width=0.195\linewidth]{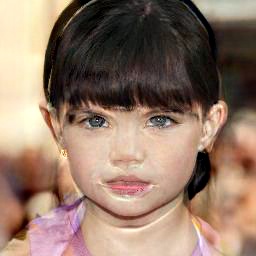}& 
     \includegraphics[width=0.195\linewidth]{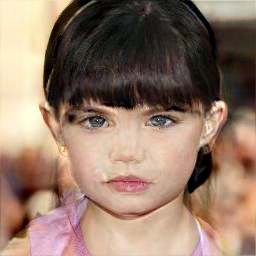}& 
     \includegraphics[width=0.195\linewidth]{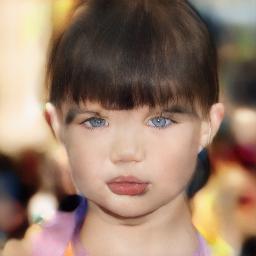}&  
     \includegraphics[width=0.195\linewidth]{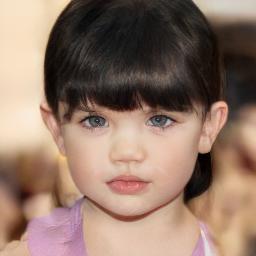}  \\
     \includegraphics[width=0.195\linewidth]{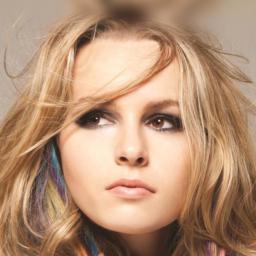}  &   
    \includegraphics[width=0.195\linewidth]{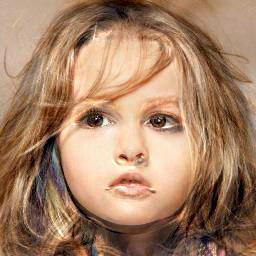}& 
     \includegraphics[width=0.195\linewidth]{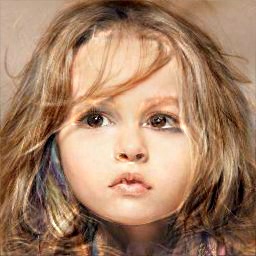}&      
     \includegraphics[width=0.195\linewidth]{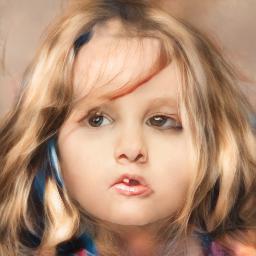} &  
     \includegraphics[width=0.195\linewidth]{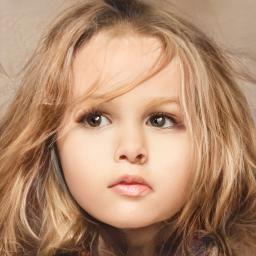}  \\
   Input& Image space& Image space &Ours&Ours \\
    &  w/o warp&   w/ warp & w/o ADA& w/ ADA 
    \end{tabular}
    \vspace{-2mm}
   \caption{Effects of ADA. We integrate the distortion map in the image and feature space respectively and show the editing results. }
   \vspace{-3mm}
    \label{fig:align}
\end{figure}

\begin{figure}[t]
    \centering 
    \footnotesize
    \begin{tabular}{@{}l@{\hspace{0.5mm}}l@{\hspace{1.2mm}}c@{\hspace{0.4mm}}c@{\hspace{0.4mm}}c@{\hspace{0.4mm}}c@{}}
     \rotatebox{90}{\hspace{0.1mm}   Original }   & \rotatebox{90}{ \hspace{0.8mm} Video }  &
    \includegraphics[width=0.225\linewidth]{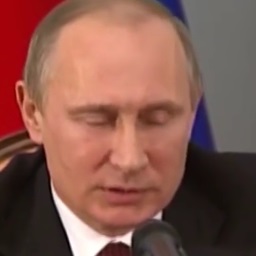} &  
    \includegraphics[width=0.225\linewidth]{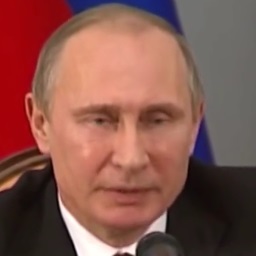}&  
    \includegraphics[width=0.225\linewidth]{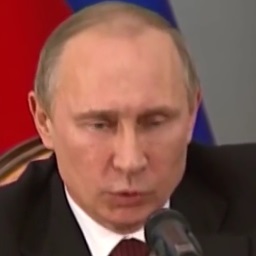}&      
    \includegraphics[width=0.225\linewidth]{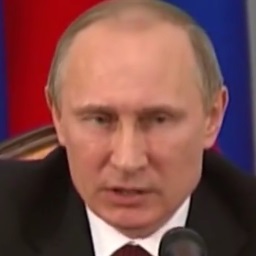}\\  
    
     \rotatebox{90}{ \hspace{2.5mm} e4e   }   & \rotatebox{90}{ (+ Smile)    }   &
    \includegraphics[width=0.225\linewidth]{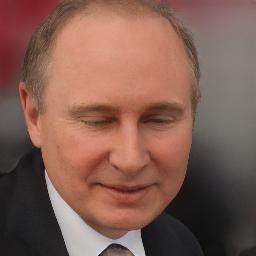} & 
    \includegraphics[width=0.225\linewidth]{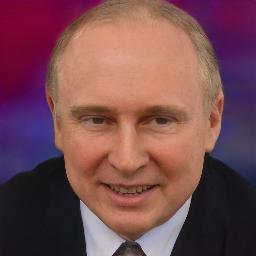}&  
    \includegraphics[width=0.225\linewidth]{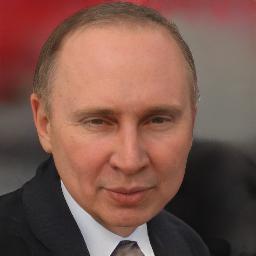}&   
    \includegraphics[width=0.225\linewidth]{cvpr/fig/abl/video/3/e4e/00002_00149.jpg}\\    

     \rotatebox{90}{ \hspace{0.6mm} Restyle  }   & \rotatebox{90}{  (+ Smile)   }   &
    \includegraphics[width=0.225\linewidth]{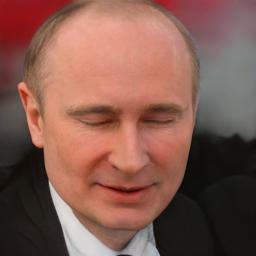}&  
    \includegraphics[width=0.225\linewidth]{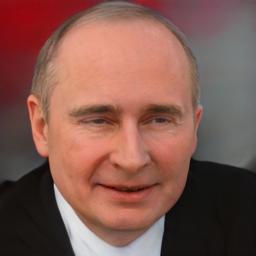}&  
    \includegraphics[width=0.225\linewidth]{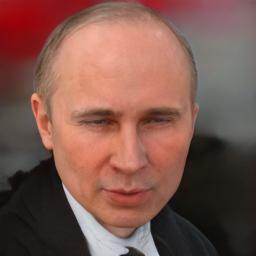}&      
    \includegraphics[width=0.225\linewidth]{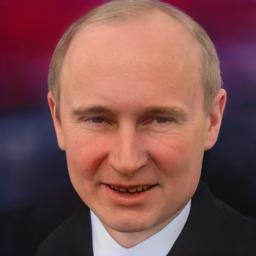}\\     

     \rotatebox{90}{ \hspace{1.7mm} Ours }   & \rotatebox{90}{ (+ Smile)   }   &
    \includegraphics[width=0.225\linewidth]{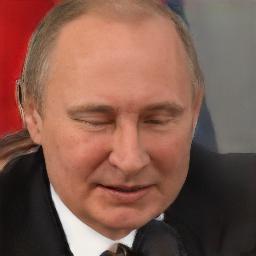} &  
    \includegraphics[width=0.225\linewidth]{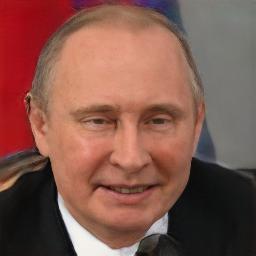}&  
    \includegraphics[width=0.225\linewidth]{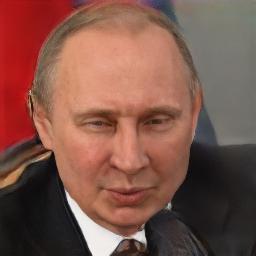}&      
    \includegraphics[width=0.225\linewidth]{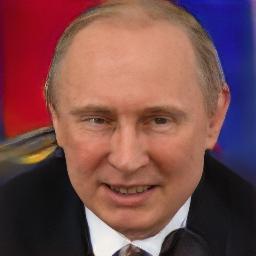}\\
    
     \rotatebox{90}{ \hspace{1.7mm} Ours   }   &  \rotatebox{90}{ \hspace{1.6mm} (Rec)   }   &
    \includegraphics[width=0.225\linewidth]{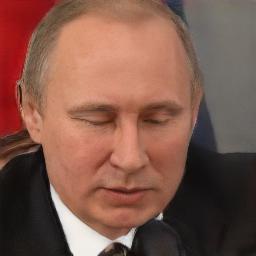} &  
    \includegraphics[width=0.225\linewidth]{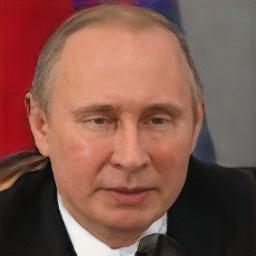}&  
    \includegraphics[width=0.225\linewidth]{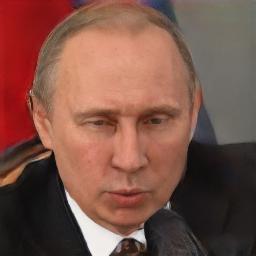}&
    \includegraphics[width=0.225\linewidth]{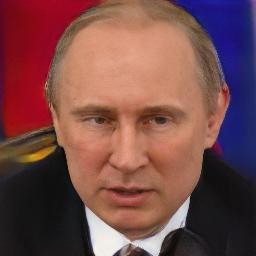}  \\
    \end{tabular}
    \vspace{-3mm}
   \caption{Inversion and editing results on a real video.}
    \label{fig:video}
    \vspace{-4mm} 
\end{figure} 

\subsection{Ablation Study} 
\label{4.3}
\subsubsection{Effect of Distortion Consultation}
\vspace{-1mm}
As discussed before, the distortion consultation inversion (DCI) scheme brings back ignored image details to complement the low-rate basic encoder, thereby achieving the high-fidelity reconstruction. To validate the effectiveness of DCI, we show our inversion results in Fig.~\ref{fig:inversion}. With the proposed distortion consultation branch, the model is more robust to occlusion and extreme poses and keeps more details in the reconstruction results. 
\vspace{-2mm}
\subsubsection{Effect of Adaptive Distortion Alignment}
\vspace{-1mm}
To analyze the effect of ADA, we show the editing results with and without ADA in Fig.~\ref{fig:align}. Without the adaptive alignment, the distortion map fails to generalize to the edited image and degrades the generated image quality. In the proposed method,  the aligned distortion map is embedded and integrated into the feature space via consultation encoding and consultation fusion. A naive alternative is to directly add the distortion map  $\Tilde{\Delta}$ to $X_o^{edit}$ in the image space with warping estimated by face landmarks~\cite{guo2020towards}. As shown in Fig.~\ref{fig:align}, performing warping and fusion in the image space also leads to obvious artifacts, where the warping is implemented by coordinates interpolation of facial landmarks.

\subsection{Application on Video Editing}
\vspace{-1mm}
Compared with image inversion and editing, the key challenge for the video counterpart is the temporal consistency of details across frames. This puts a higher demand for reconstruction fidelity since the distortion of every single image would be magnified in a video in terms of consistency and quality~\cite{Ouyang_2021_ICCV}. We show inversion and editing results on a real video~\cite{rossler2019faceforensics++}  in Fig.~\ref{fig:video}. Previous low-rate inversion approaches lack robustness to pose variation, and fail to preserve the identity of the original person and suffer notable distortion in editing results.  When the pose and viewpoint change across video frames, their results show inconsistent details and abrupt identity discrepancy. In contrast, the proposed method is more robust to cross-frame discrepancy (e.g., pose, viewpoint) and achieves higher fidelity for details preservation.  More results in \textit{mp4 format} are given in the \textbf{Supplement}.

\vspace{-1mm}
\section{Conclusion}
\vspace{-1mm}

In this work, we propose a novel GAN inversion framework that enables high-fidelity image attribute editing. With an information consultation branch, we consult the observed distortion map as a high-rate reference for lost information. This scheme enhances the basic encoder for high-quality reconstruction without compromising editability. With the adaptive distortion alignment and distortion consultation technique, our method is more robust to challenging cases such as images with occlusion and extreme viewpoints. Benefiting from the additional information of the consultation branch, the proposed method shows clear improvements in terms of image-specific details preservation (e.g., background, appearance, and illumination) for both reconstruction and editing.  The proposed framework is simple to apply, and we believe it can be easily generalized to other GAN models for future work.

\noindent\textbf{Limitations.} One limitation of the proposed method is the difficulty in handling large misalignment cases. As the augmented data used for ADA training in our experiments does not cover extreme misalignment, ADA is possibly insufficient when editing images with large viewpoint changes (see   the \textbf{Supplement} for failure cases).  

{\small
\bibliographystyle{ieee_fullname}
\bibliography{egbib}

\begin{thebibliography}{10}\itemsep=-1pt

\bibitem{abdal2019image2stylegan}
Rameen Abdal, Yipeng Qin, and Peter Wonka.
\newblock Image2stylegan: How to embed images into the stylegan latent space?
\newblock In {\em Proceedings of the IEEE/CVF International Conference on
  Computer Vision (ICCV)}, 2019.

\bibitem{abdal2020image2stylegan++}
Rameen Abdal, Yipeng Qin, and Peter Wonka.
\newblock Image2stylegan++: How to edit the embedded images?
\newblock In {\em Proceedings of the IEEE/CVF Conference on Computer Vision and
  Pattern Recognition (CVPR)}, 2020.

\bibitem{abdal2020styleflow}
Rameen Abdal, Peihao Zhu, Niloy Mitra, and Peter Wonka.
\newblock Styleflow: Attribute-conditioned exploration of stylegan-generated
  images using conditional continuous normalizing flows.
\newblock In {\em SIGGRAPH}, 2021.

\bibitem{alaluf2021restyle}
Yuval Alaluf, Or Patashnik, and Daniel Cohen-Or.
\newblock Restyle: A residual-based stylegan encoder via iterative refinement.
\newblock {\em Proceedings of the IEEE/CVF International Conference on Computer
  Vision (ICCV)}, 2021.

\bibitem{blau2019rethinking}
Yochai Blau and Tomer Michaeli.
\newblock Rethinking lossy compression: The rate-distortion-perception
  tradeoff.
\newblock In {\em International Conference on Machine Learning (ICML)}, 2019.

\bibitem{ChenPSFRGAN}
Chaofeng Chen, Xiaoming Li, Yang Lingbo, Xianhui Lin, Lei Zhang, and
  Kwan-Yee~K. Wong.
\newblock Progressive semantic-aware style transformation for blind face
  restoration.
\newblock In {\em Proceedings of the IEEE/CVF Conference on Computer Vision and
  Pattern Recognition (CVPR)}, 2021.

\bibitem{cover1999elements}
Thomas~M Cover.
\newblock {\em Elements of information theory}.
\newblock John Wiley \& Sons, 1999.

\bibitem{deng2019arcface}
Jiankang Deng, Jia Guo, Niannan Xue, and Stefanos Zafeiriou.
\newblock Arcface: Additive angular margin loss for deep face recognition.
\newblock In {\em Proceedings of the IEEE/CVF Conference on Computer Vision and
  Pattern Recognition (CVPR)}, 2019.

\bibitem{goodfellow2014generative}
Ian Goodfellow, Jean Pouget-Abadie, Mehdi Mirza, Bing Xu, David Warde-Farley,
  Sherjil Ozair, Aaron Courville, and Yoshua Bengio.
\newblock Generative adversarial nets.
\newblock In {\em Conference on Neural Information Processing Systems
  (NeurIPS)}, 2014.

\bibitem{gu2020image}
Jinjin Gu, Yujun Shen, and Bolei Zhou.
\newblock Image processing using multi-code gan prior.
\newblock In {\em Proceedings of the IEEE/CVF Conference on Computer Vision and
  Pattern Recognition (CVPR)}, 2020.

\bibitem{guan2020collaborative}
Shanyan Guan, Ying Tai, Bingbing Ni, Feida Zhu, Feiyue Huang, and Xiaokang
  Yang.
\newblock Collaborative learning for faster stylegan embedding.
\newblock {\em arXiv preprint arXiv:2007.01758}, 2020.

\bibitem{guo2020towards}
Jianzhu Guo, Xiangyu Zhu, Yang Yang, Fan Yang, Zhen Lei, and Stan~Z Li.
\newblock Towards fast, accurate and stable 3d dense face alignment.
\newblock In {\em European Conference on Computer Vision (ECCV)}, 2020.

\bibitem{harkonen2020ganspace}
Erik Harkonen, Aaron Hertzmann, Jaakko Lehtinen, and Sylvain Paris.
\newblock Ganspace: Discovering interpretable gan controls.
\newblock In {\em Conference on Neural Information Processing Systems
  (NeurIPS)}, 2020.

\bibitem{Huang2017}
Xun Huang and Serge~J. Belongie.
\newblock Arbitrary style transfer in real-time with adaptive instance
  normalization.
\newblock In {\em ICCV}, 2017.

\bibitem{huh2020transforming}
Minyoung Huh, Richard Zhang, Jun-Yan Zhu, Sylvain Paris, and Aaron Hertzmann.
\newblock Transforming and projecting images into class-conditional generative
  networks.
\newblock In {\em European Conference on Computer Vision (ECCV)}, 2020.

\bibitem{jahanian2019steerability}
Ali Jahanian, Lucy Chai, and Phillip Isola.
\newblock On the" steerability" of generative adversarial networks.
\newblock {\em The International Conference on Learning Representations
  (ICLR)}, 2020.

\bibitem{kang2021gan}
Kyoungkook Kang, Seongtae Kim, and Sunghyun Cho.
\newblock Gan inversion for out-of-range images with geometric transformations.
\newblock {\em Proceedings of the IEEE/CVF International Conference on Computer
  Vision (ICCV)}, 2021.

\bibitem{karras2018progressive}
Tero Karras, Timo Aila, Samuli Laine, and Jaakko Lehtinen.
\newblock Progressive growing of gans for improved quality, stability, and
  variation.
\newblock {\em The International Conference on Learning Representations
  (ICLR)}, 2018.

\bibitem{karras2019style}
Tero Karras, Samuli Laine, and Timo Aila.
\newblock A style-based generator architecture for generative adversarial
  networks.
\newblock In {\em Proceedings of the IEEE/CVF Conference on Computer Vision and
  Pattern Recognition (CVPR)}, 2019.

\bibitem{karras2020analyzing}
Tero Karras, Samuli Laine, Miika Aittala, Janne Hellsten, Jaakko Lehtinen, and
  Timo Aila.
\newblock Analyzing and improving the image quality of stylegan.
\newblock In {\em Proceedings of the IEEE/CVF Conference on Computer Vision and
  Pattern Recognition (CVPR)}, 2020.

\bibitem{Krause_3DRR2013}
Jonathan Krause, Michael Stark, Jia Deng, and Li Fei-Fei.
\newblock 3d object representations for fine-grained categorization.
\newblock In {\em International IEEE Workshop on 3D Representation and
  Recognition}, 2013.

\bibitem{lu2020unsupervised}
Yu-Ding Lu, Hsin-Ying Lee, Hung-Yu Tseng, and Ming-Hsuan Yang.
\newblock Unsupervised discovery of disentangled manifolds in gans.
\newblock {\em arXiv preprint arXiv:2011.11842}, 2020.

\bibitem{Ouyang_2021_ICCV}
Hao Ouyang, Tengfei Wang, and Qifeng Chen.
\newblock Internal video inpainting by implicit long-range propagation.
\newblock In {\em Proceedings of the IEEE/CVF International Conference on
  Computer Vision (ICCV)}, pages 14579--14588, October 2021.

\bibitem{plumerault2020controlling}
Antoine Plumerault, Herv{\'e}~Le Borgne, and C{\'e}line Hudelot.
\newblock Controlling generative models with continuous factors of variations.
\newblock {\em The International Conference on Learning Representations
  (ICLR)}, 2020.

\bibitem{radford2015unsupervised}
Alec Radford, Luke Metz, and Soumith Chintala.
\newblock Unsupervised representation learning with deep convolutional
  generative adversarial networks.
\newblock {\em The International Conference on Learning Representations
  (ICLR)}, 2016.

\bibitem{richardson2020encoding}
Elad Richardson, Yuval Alaluf, Or Patashnik, Yotam Nitzan, Yaniv Azar, Stav
  Shapiro, and Daniel Cohen-Or.
\newblock Encoding in style: a stylegan encoder for image-to-image translation.
\newblock {\em Proceedings of the IEEE/CVF Conference on Computer Vision and
  Pattern Recognition (CVPR)}, 2021.

\bibitem{roich2021pivotal}
Daniel Roich, Ron Mokady, Amit~H Bermano, and Daniel Cohen-Or.
\newblock Pivotal tuning for latent-based editing of real images.
\newblock {\em arXiv preprint arXiv:2106.05744}, 2021.

\bibitem{rossler2019faceforensics++}
Andreas Rossler, Davide Cozzolino, Luisa Verdoliva, Christian Riess, Justus
  Thies, and Matthias Nie{\ss}ner.
\newblock Faceforensics++: Learning to detect manipulated facial images.
\newblock In {\em Proceedings of the IEEE/CVF International Conference on
  Computer Vision (ICCV)}, 2019.

\bibitem{shannon1959coding}
Claude~E Shannon et~al.
\newblock Coding theorems for a discrete source with a fidelity criterion.
\newblock {\em IRE Nat. Conv. Rec}, 4(142-163):1, 1959.

\bibitem{shen2020interpreting}
Yujun Shen, Jinjin Gu, Xiaoou Tang, and Bolei Zhou.
\newblock Interpreting the latent space of gans for semantic face editing.
\newblock In {\em Proceedings of the IEEE/CVF Conference on Computer Vision and
  Pattern Recognition (CVPR)}, 2020.

\bibitem{shwartz2017opening}
Ravid Shwartz-Ziv and Naftali Tishby.
\newblock Opening the black box of deep neural networks via information.
\newblock {\em arXiv preprint arXiv:1703.00810}, 2017.

\bibitem{tishby2000information}
Naftali Tishby, Fernando~C Pereira, and William Bialek.
\newblock The information bottleneck method.
\newblock {\em Proceedings of Annual Allerton Conference on Communication,
  Control and Computing}, 1999.

\bibitem{tishby2015deep}
Naftali Tishby and Noga Zaslavsky.
\newblock Deep learning and the information bottleneck principle.
\newblock In {\em 2015 IEEE Information Theory Workshop (ITW)}, pages 1--5,
  2015.

\bibitem{tov2021designing}
Omer Tov, Yuval Alaluf, Yotam Nitzan, Or Patashnik, and Daniel Cohen-Or.
\newblock Designing an encoder for stylegan image manipulation.
\newblock {\em ACM Transactions on Graphics (TOG)}, 40(4):1--14, 2021.

\bibitem{voynov2020unsupervised}
Andrey Voynov and Artem Babenko.
\newblock Unsupervised discovery of interpretable directions in the gan latent
  space.
\newblock In {\em International Conference on Machine Learning (ICML)}, 2020.

\bibitem{Wang_2021_ICCV}
Tengfei Wang, Jiaxin Xie, Wenxiu Sun, Qiong Yan, and Qifeng Chen.
\newblock Dual-camera super-resolution with aligned attention modules.
\newblock In {\em Proceedings of the IEEE/CVF International Conference on
  Computer Vision (ICCV)}, pages 2001--2010, October 2021.

\bibitem{wang2021gfpgan}
Xintao Wang, Yu Li, Honglun Zhang, and Ying Shan.
\newblock Towards real-world blind face restoration with generative facial
  prior.
\newblock In {\em Proceedings of the IEEE/CVF Conference on Computer Vision and
  Pattern Recognition (CVPR)}, 2021.

\bibitem{wei2021simple}
Tianyi Wei, Dongdong Chen, Wenbo Zhou, Jing Liao, Weiming Zhang, Lu Yuan, Gang
  Hua, and Nenghai Yu.
\newblock A simple baseline for stylegan inversion.
\newblock {\em arXiv preprint arXiv:2104.07661}, 2021.

\bibitem{wu2020stylespace}
Zongze Wu, Dani Lischinski, and Eli Shechtman.
\newblock Stylespace analysis: Disentangled controls for stylegan image
  generation.
\newblock {\em Proceedings of the IEEE/CVF Conference on Computer Vision and
  Pattern Recognition (CVPR)}, 2021.

\bibitem{xia2021gan}
Weihao Xia, Yulun Zhang, Yujiu Yang, Jing-Hao Xue, Bolei Zhou, and Ming-Hsuan
  Yang.
\newblock Gan inversion: A survey.
\newblock {\em arXiv preprint arXiv:2101.05278}, 2021.

\bibitem{xu2020generative}
Yinghao Xu, Yujun Shen, Jiapeng Zhu, Ceyuan Yang, and Bolei Zhou.
\newblock Generative hierarchical features from synthesizing images.
\newblock In {\em Proceedings of the IEEE/CVF Conference on Computer Vision and
  Pattern Recognition (CVPR)}, 2021.

\bibitem{yuksel2021latentclr}
Oguz~Kaan Yuksel, Enis Simsar, Ezgi~Gülperi Er, and Pinar Yanardag.
\newblock Latentclr: A contrastive learning approach for unsupervised discovery
  of interpretable directions.
\newblock {\em arXiv preprint arXiv:2104.00820}, 2021.

\bibitem{zhang2018perceptual}
Richard Zhang, Phillip Isola, Alexei~A Efros, Eli Shechtman, and Oliver Wang.
\newblock The unreasonable effectiveness of deep features as a perceptual
  metric.
\newblock In {\em Proceedings of the IEEE/CVF Conference on Computer Vision and
  Pattern Recognition (CVPR)}, 2018.

\bibitem{zhu2020domain}
Jiapeng Zhu, Yujun Shen, Deli Zhao, and Bolei Zhou.
\newblock In-domain gan inversion for real image editing.
\newblock In {\em European Conference on Computer Vision (ECCV)}, 2020.

\bibitem{zhu2016generative}
Jun-Yan Zhu, Philipp Kr{\"a}henb{\"u}hl, Eli Shechtman, and Alexei~A Efros.
\newblock Generative visual manipulation on the natural image manifold.
\newblock In {\em European Conference on Computer Vision (ECCV)}, 2016.

\bibitem{zhu2020improved}
Peihao Zhu, Rameen Abdal, Yipeng Qin, John Femiani, and Peter Wonka.
\newblock Improved stylegan embedding: Where are the good latents?
\newblock {\em arXiv preprint arXiv:2012.09036}, 2020.

\bibitem{zhuang2021enjoy}
Peiye Zhuang, Oluwasanmi Koyejo, and Alexander~G Schwing.
\newblock Enjoy your editing: Controllable gans for image editing via latent
  space navigation.
\newblock {\em The International Conference on Learning Representations
  (ICLR)}, 2021.

\end{thebibliography}
}

\end{document}